\documentclass{article}

\usepackage{microtype}
\usepackage{graphicx}
\usepackage{subfigure}
\usepackage{booktabs} 

\usepackage{hyperref}

\usepackage{cite}
\usepackage{amsmath,amssymb,amsfonts}
\usepackage{graphicx}
\usepackage{textcomp}
\usepackage{xcolor}
\usepackage{fancyhdr}
\usepackage{multicol}
\usepackage{multirow}

\usepackage{soul}
\usepackage[utf8]{inputenc}
\usepackage{graphicx}
\usepackage{booktabs}

\usepackage{multicol}

\usepackage{graphicx}
\usepackage{booktabs} 

\usepackage[ruled,noend]{algorithm2e}

\SetCommentSty{mycommfont}

\usepackage{here}

\usepackage{amsmath,amssymb,amsfonts,amsbsy,amsfonts,latexsym}
\usepackage{multirow}
\usepackage{makecell}
\usepackage[labelfont=bf,textfont=it,belowskip=0pt,aboveskip=5pt,tableposition=top]{caption}
\usepackage{xcolor}
\usepackage{colortbl}

\definecolor{colorA}{RGB}{189,201,225}
\definecolor{colorB}{RGB}{103,169,207}
\definecolor{colorC}{RGB}{ 28,144,153}
\definecolor{colorD}{RGB}{  1,108, 89}

\newcolumntype{R}{>{\columncolor{gray!40}}r}
\newcolumntype{L}{>{\columncolor{gray!40}}l}
\newcolumntype{C}{>{\columncolor{gray!40}}c}

\usepackage{tabularx,colortbl,xcolor}
\usepackage{multirow}
\usepackage[normalem]{ulem}
\useunder{\uline}{\ul}{}

\usepackage{enumitem}

\usepackage{xparse}

\DeclareGraphicsExtensions{.pdf,.png}

\SetKwInput{KwInput}{Input}

\usepackage{longtable}
\usepackage{pgfplots}
\usepackage{outlines}

\newcommand\fref{Fig.~\ref}
\newcommand\tref{Tab.~\ref}
\newcommand\sref{Sec.~\ref}

\usepackage[many]{tcolorbox}
\usepackage{caption}
\usepackage{subcaption}
\usepackage{graphbox} 

\tcbset{
    sharp corners,
    colback = white,
    before skip = 0.2cm,    
    after skip = 0.5cm      
}                           

\definecolor{main}{HTML}{4472C4}    
\definecolor{sub}{HTML}{EBF4FF}     

\newtcolorbox{boxA}{
    enhanced, breakable,
    boxrule = 0pt,
    colback = sub,
    borderline west = {2pt}{0pt}{main}, 
    borderline east = {2pt}{0pt}{main}, 
}

\newcommand\hc{ \rowcolor{teal!15}}

\newcommand{\graycell}[1]{\cellcolor{teal!15}#1}

\usepackage[accepted]{icml2024}

\usepackage{amsmath}
\usepackage{amssymb}
\usepackage{mathtools}
\usepackage{amsthm}
\usepackage[defaultlines=2,all]{nowidow}

\usepackage[capitalize,noabbrev]{cleveref}

\theoremstyle{plain}

\theoremstyle{definition}

\theoremstyle{remark}

\usepackage{subcaption}
\usepackage[textsize=tiny]{todonotes}
\usepackage{placeins}

\usepackage{amsmath}

\newcommand{\OURS}{\texttt{LLMCompiler}\xspace}
\newcommand{\BENCH}{{ParallelQA}\xspace}

\begin{document}

\twocolumn[
\icmltitle{An LLM Compiler for Parallel Function Calling}



\icmlsetsymbol{equal}{*}

\begin{icmlauthorlist}
\icmlauthor{Sehoon Kim}{equal,berkeley}
\icmlauthor{Suhong Moon}{equal,berkeley}
\icmlauthor{Ryan Tabrizi}{berkeley}
\icmlauthor{Nicholas Lee}{berkeley}
\icmlauthor{Michael W. Mahoney}{berkeley,icsi,lbnl}
\\
\icmlauthor{Kurt Keutzer}{berkeley}
\icmlauthor{Amir Gholami}{berkeley,icsi}

\end{icmlauthorlist}

\icmlaffiliation{berkeley}{UC Berkeley}
\icmlaffiliation{icsi}{ICSI}
\icmlaffiliation{lbnl}{LBNL}

\icmlcorrespondingauthor{Amir Gholami}{amirgh@berkeley.edu}

\icmlkeywords{Machine Learning, ICML}

\vskip 0.3in
]



\printAffiliationsAndNotice{\icmlEqualContribution} 
\begin{abstract}
The reasoning capabilities of the recent LLMs enable them to execute external function calls to overcome their inherent limitations, such as knowledge cutoffs, poor arithmetic skills, or lack of access to private data.
This development has allowed LLMs to select and coordinate multiple functions based on the context to tackle more complex problems.
However, current methods for function calling often require
sequential reasoning and acting for each function which can result
in high latency, cost, and sometimes inaccurate behavior.
To address this, we introduce \OURS, which executes functions in parallel to efficiently orchestrate multiple function calls.
Drawing inspiration from the principles of classical compilers, \OURS enables parallel function calling with three components:
(i) a Function Calling Planner, formulating execution plans for function calling;
(ii) a Task Fetching Unit, dispatching function calling tasks;
and (iii) an Executor, executing these tasks in parallel.
\OURS automatically generates an optimized orchestration for the function calls and can be used with both open-source and closed-source models.
We have benchmarked \OURS on a range of tasks with different
patterns of function calling.
We observe consistent latency speedup of up to $3.7\times$, cost
savings of up to $6.7\times$, 
and accuracy improvement of up to ${\sim}9\%$ compared to ReAct.
Our code is available at \href{https://github.com/SqueezeAILab/LLMCompiler}{https://github.com/SqueezeAILab/LLMCompiler}.
\end{abstract}

\vspace{-6mm}
\section{Introduction}
\vspace{-1mm}
\begin{figure*}[!t]
    \centering
    \includegraphics[width=0.99\linewidth]{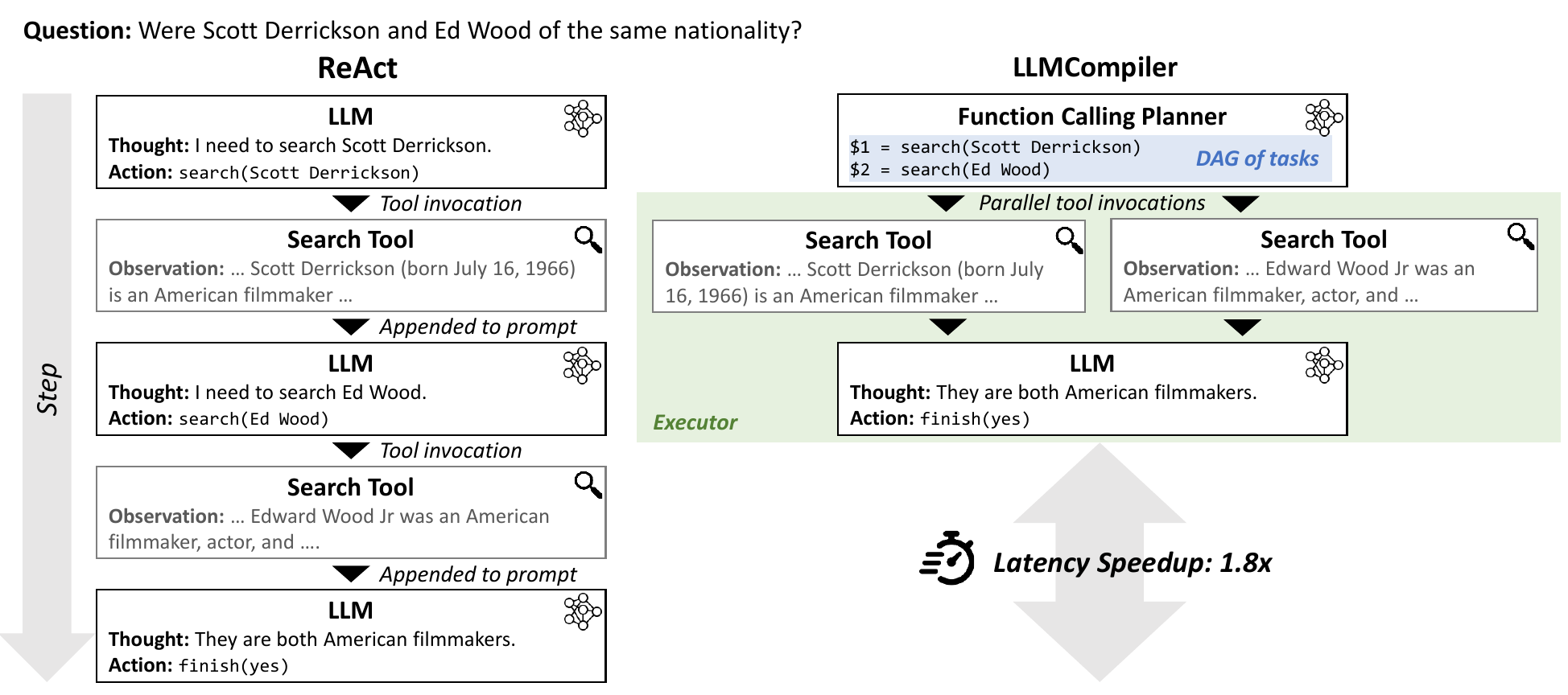}
        \vspace*{-4mm}
    \caption{
    An illustration of the runtime dynamics of \OURS, in comparison with ReAct~\cite{yao2022react}, given a sample question from the HotpotQA benchmark~\cite{yang2018hotpotqa}. 
    In \OURS (Right), the Planner first decomposes the query into several tasks with inter-dependencies.
    The Executor then executes multiple tasks in parallel, respecting their dependencies. 
    Finally, \OURS joins all observations from the tool executions to produce the final response. 
    In contrast, sequential tool execution of the existing frameworks like ReAct (Left) leads to longer execution latency. 
    In this example, \OURS attains a latency speedup of 1.8$\times$ on the HotpotQA benchmark.
    While a 2-way parallelizable question from HotpotQA is presented here for the sake of simple visual illustration, 
    \OURS is capable of managing tasks with more complex dependency patterns (Fig.~\ref{fig:system_overview} and Sec.~\ref{sec:results}).
    }
    \vspace{-4mm}
    \label{fig:teaser}
\end{figure*}


Recent advances in the reasoning capability of Large Language Models (LLMs)
have expanded the applicability of LLMs beyond content generation to solving complex
problems~\cite{wei2022cot,kojima2023large,yao2023tree,besta2023got,wang2023selfconsistency,zhou2023leasttomost,chen2023program,yang2022language,gao2022pal};
and recent works have also shown how this reasoning capability can be helpful
in improving accuracy for solving complex and logical tasks.
The reasoning capability has also allowed function (i.e., tool) calling capability, where LLMs can invoke provided functions and use
the function outputs to help complete their tasks. 
These functions range from a simple calculator that can invoke arithmetic operations to more complex LLM-based functions.

The ability of LLMs to integrate various tools and function calls
could enable a fundamental shift in how we develop LLM-based software.
However, this brings up an important challenge: \textit{what is the most effective approach to incorporate multiple function calls?}
A notable approach has been introduced in ReAct~\cite{yao2022react}, where the LLM calls a function, analyzes the outcomes, and then reasons about the next action, which involves a subsequent function call.
For a simple example illustrated in~\fref{fig:teaser} (Left), where the LLM is asked if Scott Derrickson and Ed Wood have the same nationality,
ReAct initially analyzes the query and decides to use a search tool to search for Scott Derrickson. 
The result of this search (i.e., observation) is then concatenated back to the original prompt for the LLM to reason about the next action, 
which invokes another search tool to gather information about Ed Wood.

ReAct has been a pioneering work in enabling function calling, and it has been integrated into several frameworks~\cite{langchain,Liu_LlamaIndex_2022}. 
However, scaling this approach for more complex applications requires considerable optimizations. 
This is due to the sequential nature of ReAct, 
where it executes function calls and reasons about their observations one after the other.
This approach, along with the agent systems that extend ReAct~\cite{khot2023decomposed,yao2023tree,qin2023toolllm,ruan2023toolemu,sumers2023cognitive}, may lead to inefficiencies in latency and cost, due to the sequential function calling and repetitive LLM invocations for each reasoning and action step.
 Furthermore, while dynamic reasoning about the observations has benefits in certain cases, 
 concatenating the outcomes of intermediate function calls could disrupt the LLM's execution flow,
potentially reducing accuracy~\cite{xu2023rewoo}.
Common failure cases include repetitive invocation of the same function, which is also highlighted in the original paper~\cite{yao2022react}, and early stopping based on the partial intermediate results, as will be further discussed in Sec.~\ref{subsec:parallelizable_workloads} and Appendix~\ref{subsec:accuracy_analysis}.

To address this challenge, we draw inspiration from classical compilers, 
where optimizing instruction executions in traditional programming languages has been extensively explored.
A key optimization technique in compilers involves 
identifying instructions that can be executed in parallel and effectively managing their dependencies. 
Similarly, one can envision a {compiler}, tailored for LLM function calling, which can efficiently orchestrate various function calls and their dependencies. 
This shares a similar philosophy with the recent studies that align LLMs with computer systems~\cite{andrej, packer2023memgpt}.
To this end, we introduce \OURS, a novel framework that enables parallel multi-tool execution of LLMs across different models and workloads. To the best of our knowledge, \OURS is the first framework to optimize the orchestration of LLM function calling that can not only improve latency and cost, but
also accuracy, by minimizing interference from the outputs of intermediate function calls.
In more detail, we make the following contributions:

\vspace{-2mm}
\begin{itemize}[leftmargin=3mm]
    \item We introduce \OURS, an LLM compiler that optimizes the parallel function calling
    performance of LLMs. 
    At a high level, this is achieved by introducing three key components:
    (i) a Function Calling Planner (Sec.~\ref{subsec:planner}) that identifies an execution flow;
    (ii) a Task Fetching Unit (Sec.~\ref{subsec:tfu}) that dispatches the function calls in parallel;
    (iii) an Executor (Sec.~\ref{subsec:executor}) that executes the dispatched tasks using the associated functions. 
    \vspace{-2mm}

    \item    We evaluate \OURS on \textit{embarrassingly parallel}
    patterns using HotpotQA \cite{yang2018hotpotqa} and Movie Recommendation~\cite{srivastava2023beyond}, where we observe $1.80\times$/$3.74\times$ speedup and
    $3.37\times$/$6.73\times$ cost reduction compared to ReAct (Sec.~\ref{subsec:parallelizable_workloads}).
    \vspace{-2mm}

    \item To test the performance on more complex patterns, we introduce a new benchmark called
    \BENCH which includes various non-trival function calling patterns. 
    We show up to $2.27\times$ speedup, $4.65\times$ cost
    reduction, and $9\%$ improved accuracy compared to ReAct (Sec.~\ref{subsed:multi-tool}).
    \vspace{-2mm}

    \item We evaluate \OURS's capability in dynamic replanning, which is achieved through a feedback loop from the Executor back to our Function Calling Planner.
    For the Game of 24~\cite{yao2023tree}, which requires repeated replanning based on the intermediate results, \OURS demonstrates a 2$\times$ speedup compared to Tree-of-Thoughts (Sec.~\ref{subsec:replanning-heavy}).
    \vspace{-2mm}

    \item We show that \OURS can explore the interactive decision-making environment effectively and efficiently. On WebShop, \OURS achieves up to $101.7\times$ speedup and $25.7\%$ improved success rate compared to the baselines (Sec.~\ref{subsec:interactive_decision_making_tasks}).
        \vspace{-2mm}

\end{itemize}

\section{Related Work}
\label{sxn:related_work_x}

\subsection{Latency Optimization in LLMs}
Various studies have focused on optimizing model design~\cite{kim2023squeezellm,frantar-gptq,lin2023awq,dettmers2023spqr,kwon2022fast,frantar2023sparsegpt,kim2023bild,chen2023accelerating,leviathan2023fast}  and systems~\cite{kwon2023vllm,yu2022orca,trtllm,tgi} for efficient LLM inference. Optimizations at the application level, however, are less explored. This is critical from a practical point of view for situations involving black-box LLM models and services where modifications to the models and the underlying inference pipeline are highly restricted.

Skeleton-of-Thought~\cite{ning2023skeletonofthought} recently proposed to reduce latency through application-level parallel decoding.
This method involves a two-step process of an initial skeleton generation phase, followed by parallel execution of skeleton items.
However, it is primarily designed for embarrassingly parallel workloads and does not support problems that have inherently interdependent tasks, as it assumes no dependencies between skeleton tasks.
This limits its applicability in complex scenarios such as coding~\cite{chen2021codex,madaan2023selfrefine,hendrycksapps2021,austin2021program} or math~\cite{hendrycksmath2021,hendryckstest2021} problems, as also stated in the paper~\cite{ning2023skeletonofthought}.
\OURS addresses this by translating an input query into a series of tasks with inter-dependencies, thereby expanding the spectrum of problems it can handle.

Concurrently to our work, OpenAI has recently introduced a parallel function calling feature in their 1106 release, 
enhancing user query processing through the simultaneous generation of multiple function calls~\cite{openai_devday}.
Despite its potential for reducing LLM execution time, this feature has certain limitations, as it is exclusively available for OpenAI's proprietary models.
However, there is a growing demand for using open-source models driven by the increasing number of open-source LLMs
as well as parameter-efficient training techniques~\cite{lester2021power,hu2022lora,pmlr-v97-houlsby19a}
for finetuning and customization.
\OURS enables efficient parallel function calling for open-source models, and
also, as we will show later in Sec.~\ref{sec:results}, it can potentially achieve better latency and cost.

\begin{figure*}[!t]
\vspace{-1mm}
    \centering
    \includegraphics[width=0.99\linewidth]{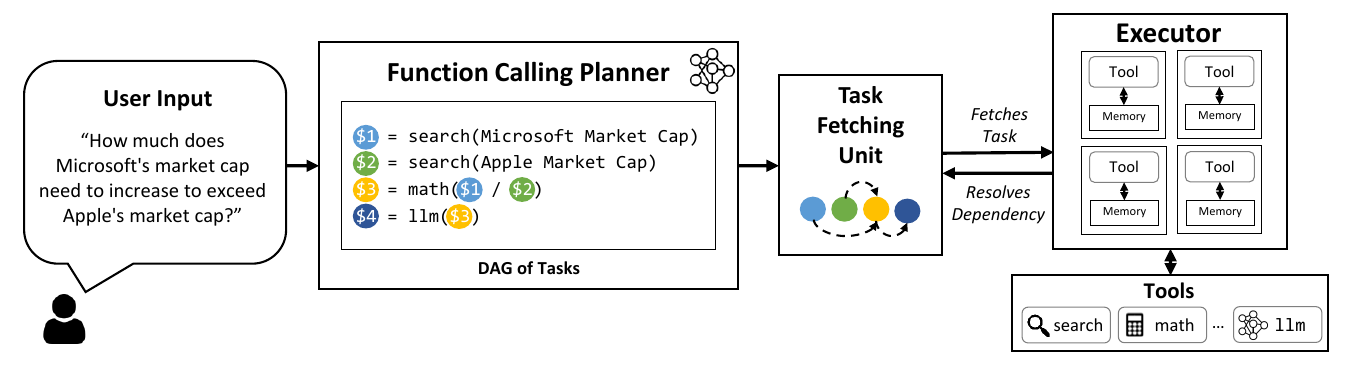}
        \vspace*{-4mm}
    \caption{Overview of the \OURS framework.
    The Function Calling Planner generates a DAG of tasks with their inter-dependencies. 
    These tasks are then dispatched by the Task Fetching Unit to the Executor in parallel based on their dependencies.
    In this example, Task \texttt{\$1} and \texttt{\$2} are fetched together for parallel execution of two independent search tasks.
    After each task is performed, the results  are forwarded back to the Task Fetching Unit to unblock the dependent tasks after replacing their placeholder variables (e.g., the variable \texttt{\$1} and \texttt{\$2} in Task \texttt{\$3}) with actual values. 
    Once all tasks have been executed, the final answer is delivered to the user.
    }
    \vspace{-1mm}
    \label{fig:system_overview}
\end{figure*}

\subsection{Plan and Solve Strategy}
Several studies~\cite{Wolfson2020Break, patel-etal-2022-question, press2023measuring, zhou2023leasttomost, hao2023reasoning} have explored prompting methods of breaking down complex queries into various levels of detail to solve them, 
thereby improving LLM's performance in reasoning tasks. 
Specifically, Decomposed Prompting~\cite{khot2023decomposed} tackles complex tasks by decomposing them into simpler sub-tasks, each optimized through LLMs with dedicated prompts. 
Step-Back Prompting~\cite{zheng2023step} enables LLMs to abstract high-level concepts from details to enhance reasoning abilities across various tasks. 
Plan-and-Solve Prompting~\cite{wang2023plan} segments multi-step reasoning tasks into subtasks to minimize errors and improve task accuracy without manual prompting.
However, these methods primarily focus on improving the accuracy of reasoning benchmarks. In contrast, \OURS uses a planner to identify parallelizable patterns within queries, aiming to reduce latency while maintaining accuracy.

In addition to the aforementioned works, TPTU~\cite{ruan2023tptu}, HuggingGPT~\cite{shen2023hugginggpt}, and ViperGPT~\cite{suris2023vipergpt} have introduced end-to-end plan-and-solve frameworks. 
 \OURS sets itself apart by providing a general framework that enables efficient and accurate function calling in a broader range of problems.
This stems from \OURS's capabilities in (i) planning and replanning; (ii) parallel execution; and (iii) addressing a wider range of problem domains, which will be discussed in more detail in Appendix~\ref{appendix:baseline-comparison}.

Another notable work is ReWOO~\cite{xu2023rewoo} which 
employs a planner 
to separate the reasoning process from the execution and observation phases to
decrease token usage and cost as compared to ReAct.
Our approach is different from ReWOO in multiple aspects. First, \OURS allows
parallel function calling which can reduce latency as well as cost.
Second, \OURS supports dynamic replanning which is important for problems whose execution flow
cannot be determined statically in the beginning (Sec.~\ref{subsec:replanning-heavy}).

\subsection{Tool-Augmented LLMs}

The enhanced reasoning capability of LLMs has enabled them to invoke user-provided functions and use their outputs to effectively complete tasks. 
Detailed exploration of this subject is provided in the Appendix~\ref{appendix:tool-augmented-llms}.

\section{Methodology}
\label{sec:method}

To illustrate the components of \OURS, we use a simple 2-way parallel example in~\fref{fig:system_overview}.
To answer  ``How much does Microsoft's market cap need to increase to exceed Apple's market cap?,'' the LLM first needs to conduct web searches for both companies' market caps, followed by a division operation. 
While the existing frameworks, including ReAct, perform these tasks sequentially, it is evident that they can be executed in parallel.
The key question is how to automatically determine which tasks are parallelizable and which are interdependent, so we can orchestrate the execution of the different tasks accordingly.
\OURS accomplishes this through a system that consists of the following three components:
a Function Calling Planner (\sref{subsec:planner}) that generates a sequence of tasks and their dependencies;
a Task Fetching Unit (\sref{subsec:tfu}) that replaces arguments based on intermediate results and fetches the tasks;
and an Executor (\sref{subsec:executor}) that executes the tasks with associated tools.
To use \OURS, users are only required to provide tool definitions, and optional in-context examples for the Planner, as will be further discussed in Sec.~\ref{subsec:user_info}.

\subsection{Function Calling Planner}

\label{subsec:planner}
The Function Calling Planner is responsible for generating a sequence of tasks to
be executed along with any dependency among them. For instance, Tasks \texttt{\$1} and \texttt{\$2} in~\fref{fig:system_overview} are two independent searches that can be performed in parallel.
However, Task \texttt{\$3} has a dependency on the outcomes of the first and second searches.
Therefore, the Planner's role is to automatically identify the necessary tasks, their input arguments, as well as their inter-dependencies using the sophisticated reasoning capability of LLMs, 
essentially forming a directed acyclic graph of task dependencies.
If a task is dependent on a preceding task, it incorporates a placeholder variable, such as \texttt{\$1} in Task 3 of~\fref{fig:system_overview}, which will later be substituted with the actual output from the preceding task (Sec.~\ref{subsec:tfu}).

The Planner in \OURS leverages LLMs' reasoning capability to decompose tasks from natural language inputs. 
To achieve this, the Planner LLM incorporates a pre-defined prompt that guides it on how to create dependency graphs and to ensure correct syntax (see Appendix~\ref{appendix:planner-prompt} for details).
Besides this, users also need to supply tool definitions and optional in-context examples for the Planner.
These examples provide detailed demonstrations of task decomposition specific to a problem, helping the Planner to better understand the rules. 
Further details on user-supplied information for \OURS are elaborated in Sec.~\ref{subsec:user_info}. 
In Sec.~\ref{sec:streaming}, we introduce an additional optimization for the Planner that streams tasks as soon as they are created, instead of waiting to complete the entire planning process.

\subsection{Task Fetching Unit}
\label{subsec:tfu}
The {Task Fetching Unit}, inspired by the instruction fetching units in modern computer architectures,
fetches tasks to the Executor as soon as they are ready for (parallel) execution based on a greedy policy.
Another key functionality is to replace variables with the actual outputs from preceding tasks, which were initially set as placeholders by the Planner.
For the example in~\fref{fig:system_overview}, the variable \texttt{\$1} and \texttt{\$2} in Task \texttt{\$3} would be replaced with the actual market cap of Microsoft and Apple.
This can be implemented with a simple fetching and queuing mechanism 
without a dedicated LLM.

\subsection{Executor}
\label{subsec:executor}
The {Executor} asynchronously executes tasks fetched from the Task Fetching Unit.
As the Task Fetching Unit guarantees that all the tasks dispatched to the Executor are independent, it can simply execute them concurrently.
The Executor is equipped with user-provided tools, and it delegates the task to the associated tool. 
These tools can be simple functions like a calculator, Wikipedia search, or API calls,
or they can even be LLM agents that are tailored for a specific task.
As depicted in the Executor block of~\fref{fig:system_overview}, 
each task has dedicated memory to store its intermediate outcomes,
 similar to what typical sequential frameworks do when aggregating observations as a single prompt~\cite{yao2022react}.
Upon completion of the task, 
the final results are forwarded as input to the tasks dependent on them.

\subsection{Dynamic Replanning}
In various applications, the execution graph may need to adapt based on intermediate results
that are a priori unknown.
An analogy in programming is branching, where the path of execution is determined only during runtime, depending on which branch conditions are satisfied.
Such dynamic execution patterns can also appear with LLM function calling. 
For simple branching (e.g., if-else statements) one could statically compile
the execution flow and choose the right dynamically based on the intermediate results.
However, for more complex branching it may be better to do a recompilation or replanning
based on the intermediate results.

When replanning, the intermediate results are sent back from the Executor to the Function Calling Planner which then generates a new set of tasks with their associated dependencies.
These tasks are then sent to the Task Fetching Unit and subsequently to the Executor. 
This cycle continues until the desired final result is achieved and can be delivered to the user.
We show an example use case of this in Sec.~\ref{subsec:replanning-heavy} for
solving the Game of 24 using the Tree-of-Thoughts approach.


\section{\OURS Details}
\subsection{User-Supplied Information}
\label{subsec:user_info}
\OURS requires two inputs from the user:
\vspace{-2mm}
\begin{enumerate}[leftmargin=5mm]
\item \textbf{Tool Definitions}:
Users need to specify the tools that LLMs can use, including their descriptions and argument specifications. 
This is essentially the same requirement as other frameworks like ReAct and OpenAI function calling.
\vspace{-2mm}

\item \textbf{In-context Examples for the Planner}: 
Optionally, users can provide \OURS with examples of how the Planner should behave.
For instance, in the case of~\fref{fig:system_overview}, users may provide examples illustrating expected inter-task dependencies for certain queries.
These examples can assist the Planner LLM understand how to use various tools and generate the appropriate dependency graph for incoming inputs in the correct format.
In Appendix~\ref{appendix:prompts}, we include the examples that we used in our evaluations.
\end{enumerate}

\subsection{Streamed Planner}
\label{sec:streaming}
The Planner may incur a non-trivial overhead for user queries that involve a lot of tasks
as it blocks the Task Fetching Unit and the Executor, which must wait for the Planner output before initiating their processes.
However, analogous to instruction pipelining in modern computer systems, this can be mitigated by enabling the Planner to asynchronously stream
the dependency graph, thereby allowing each task to be immediately processed by the Executor as soon as its dependencies are all resolved.
In Table~\ref{table:streaming}, we present a latency comparison of \OURS with and without the streaming mechanism across different benchmarks. 
The results demonstrate consistent latency improvements with streaming.
Particularly, in the \BENCH benchmark, the streaming feature leads to a latency gain of up to 1.3$\times$. 
This is attributed to the {math} tool's longer execution time for \BENCH, which can effectively hide the Planner's latency in generating subsequent tasks, unlike the shorter execution times of the \texttt{search} tool used in HotpotQA and Movie Recommendation.

\begin{figure*}[!t]
    \centering
    \includegraphics[width=\linewidth]{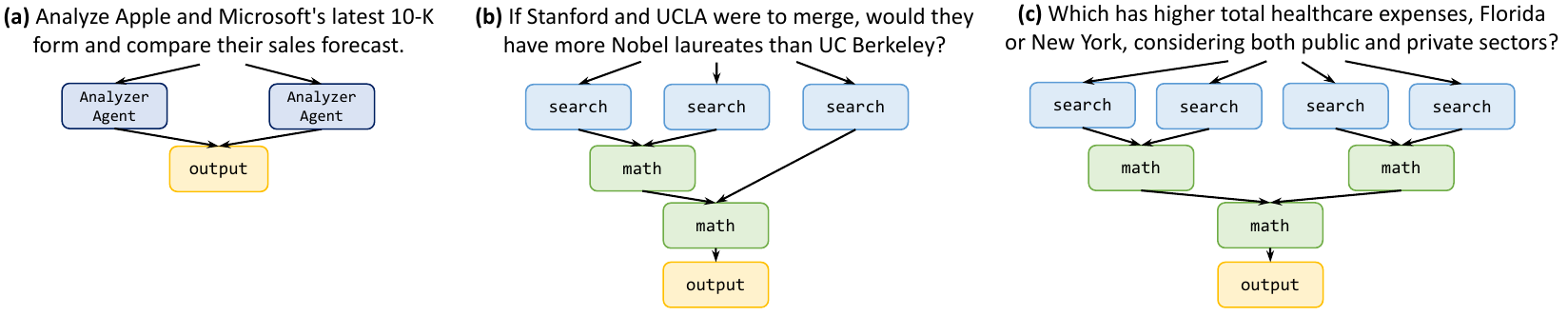}
        \vspace*{-5mm}
    \caption{Examples of questions with different function calling patterns and their dependency graphs. 
    HotpotQA and Movie Recommendation datasets exhibit pattern (a),
    and
    ParallelQA dataset exhibits patterns (b) and (c), among other patterns.
    In (a), we need to analyze each company's latest 10-K. 
    In (b), we need three searches for each school, followed by one addition and one comparison operation.
    In (c), we need to search for each state's annual healthcare spending in each sector, sum each state's spending, and then perform a comparison.
    } 
    \label{fig:parallel-patterns}
\end{figure*}

\section{Results}
\label{sec:results}
In this section, we evaluate \OURS using a variety of models and problem types. 
We use both the proprietary GPT models and the open-source LLaMA-2 model, with the latter demonstrating \OURS's capability in enabling parallel function calling in open-source models.
Furthermore, there are various types of parallel function calling patterns that can
be addressed with LLMs. This ranges from embarrassingly parallel patterns, where all tasks
can be executed in parallel without any dependencies between them, to more complex dependency patterns, as illustrated in ~\fref{fig:parallel-patterns}.
Importantly, we also assess \OURS on the Game of 24 benchmark, which involves dynamic replanning based on intermediate results, 
highlighting its adaptability to dynamic dependency graphs.
Finally, we apply \OURS to the WebShop benchmark to showcase its potential in decision-making tasks.
Overall, we start presenting results for simple execution patterns, and then we move
to more complex ones.

\begin{table*}[!t]
\vspace{-3mm}
\caption{Accuracy and latency comparison of \OURS compared to the baseline on different benchmarks, including HotpotQA, Movie Recommendation, our custom dataset named \BENCH, and the Game of 24. For HotpotQA and Movie Recommendation, we frequently observe looping and early stopping (Sec.~\ref{subsec:parallelizable_workloads}). 
To minimize these behaviors as much as possible, we incorporated ReAct-specific prompting which we denote as ReAct$^\dagger$.
ReAct (without $^\dagger$) indicates the original results without this prompting. We do not include the latency for the original ReAct since looping and early stopping make precise latency measurement difficult.
}
\vspace{-1mm}
\begin{center}
\scriptsize{
\setlength{\tabcolsep}{6pt}{
\begin{tabular}{c|c|c|cc|c|cc}
\toprule
  \multirow{2}{*}{\raisebox{-0.5ex}{\textbf{Benchmark}}} & \multirow{2}{*}{\raisebox{-0.5ex}{\textbf{Method}}} & \multicolumn{3}{c|}{\textbf{GPT (Closed-source)}}  & \multicolumn{3}{c}{\textbf{LLaMA-2 70B (Open-source)}}  \\
   \cmidrule{3-8}
& & Accuracy (\%) & Latency (s) & Speedup  & Accuracy (\%) & Latency (s) & Speedup  \\
\midrule
 \multirow{4}{*}{HotpotQA}  & ReAct & 61.52 & - & - & 54.74  & - & -\\
 & ReAct$^\dagger$ & 62.47 & 7.12 & 1.00$\times$ &  54.40 & 13.44 & 1.00$\times$\\ 
   & OAI  Parallel Function & 62.05 & 4.42 & 1.61$\times$ & -  &  - &  -\\ 
  & \graycell{\OURS} & \graycell{62.00} & \graycell{\textbf{3.95}} & \graycell{\textbf{1.80$\times$}} & \graycell{57.83} & \graycell{\textbf{9.58}} & \graycell{\textbf{1.40$\times$}} \\
\midrule
\multirow{4}{*}{Movie Rec.}  & ReAct & 68.60 & - & - & 70.00  & - & -\\ 
& ReAct$^\dagger$ & 72.47 & 20.47 & 1.00$\times$ & 70.60 & 33.37 & 1.00$\times$\\ 

  & OAI  Parallel Function & 77.00 &  7.42 & 2.76$\times$  & -  &  - &  -\\ 
 &   \graycell{\OURS} & \graycell{77.13} & \graycell{\textbf{5.47}} & \graycell{\textbf{3.74$\times$}} & \graycell{77.80} & \graycell{\textbf{11.83}} & \graycell{\textbf{2.82$\times$}} \\

\midrule
 \multirow{3}{*}{\BENCH} & ReAct & 89.09 & 35.90 & 1.00$\times$& 59.59 & 15.47 & 1.00$\times$\\ 
  &  OAI Parallel Function & 87.32  &  19.29 &  1.86$\times$  & -  &  - &  -\\  
&  \graycell{\OURS} & \graycell{89.38} & \graycell{\textbf{16.69}} & \graycell{\textbf{2.15$\times$}} & \graycell{68.14} & \graycell{\textbf{26.20}} & \graycell{\textbf{2.27$\times$}}  \\
\midrule
 \multirow{2}{*}{Game of 24} & Tree-of-Thoughts & 74.00 & 241.2 & 1.00$\times$ & 30.00 & 952.06 & 1.00$\times$ \\ 
  & \graycell{\OURS} & \graycell{75.33} & \graycell{\textbf{83.6}} & \graycell{\textbf{2.89$\times$}} & \graycell{32.00} & \graycell{\textbf{456.02}} & \graycell{\textbf{2.09$\times$}}\\
  
\bottomrule
\end{tabular}
}
}
\end{center}
\vspace{-4mm}
\label{table:benchmark}
\end{table*}

\begin{table}[!t]
\vspace{-2mm}
\caption{Input and output token consumption as well as the estimated cost on HotpotQA, Movie Recommendation, and our custom dataset named \BENCH.
The cost is computed based on the pricing table of the GPT models used for each benchmark.}
\vspace{-1mm}
\begin{center}
\scriptsize{
\setlength{\tabcolsep}{7pt}{
\begin{tabular}{c|c|cc|cc}
\toprule

 \multirow{2}{*}{Benchmark} &  \multirow{2}{*}{Method} &  \multicolumn{2}{c|}{Tokens} & Cost & Cost\\
 & & In. & Out. &  (\$/1k) &  Red. \\
\midrule
 \multirow{3}{*}{HotpotQA} & ReAct & 2900 & 120 & 5.00 & 1.00$\times$\\ 
   & OAI Para. Func. &  2500 & 63 & 2.66 & 1.87$\times$\\ 
   & \graycell{\OURS} &  \graycell{1300} & \graycell{80} & \graycell{\textbf{1.47}} & \graycell{\textbf{3.37$\times$}}\\
\midrule
\multirow{3}{*}{Movie Rec.} & ReAct & 20000 & 230 & 20.46 & 1.00$\times$\\ 
   & OAI Para. Func. &  5800 & 160 & 6.14 & 3.33$\times$\\ 
   & \graycell{\OURS} &  \graycell{2800} & \graycell{115} & \graycell{\textbf{3.04}} & \graycell{\textbf{6.73$\times$}}\\
 
\midrule
 \multirow{3}{*}{\BENCH} & ReAct & 46000 & 470 & 480 & 1.00$\times$ \\ 
   & OAI Para. Func. & 25000 & 370 & 260 & 1.81$\times$\\ 
   & \graycell{\OURS}  & \graycell{9200} & \graycell{340} & \graycell{\textbf{103}} & \graycell{\textbf{4.65$\times$}} \\
\bottomrule
\end{tabular}
}
}
\end{center}
\vspace{-6mm}
\label{table:cost}
\end{table}


\subsection{Embarrassingly Parallel Function Calling}
\label{subsec:parallelizable_workloads}
The simplest scenario involves an LLM using a tool repeatedly for independent tasks such as conducting parallel searches or analyses to gather information on different topics, like the pattern depicted in~\fref{fig:parallel-patterns} (a).
While these tasks are independent of each other and can be executed in parallel,
ReAct, along with other LLM solutions as they stand, would need to run sequentially.
This leads to increased latency and token consumption due to its frequent LLM invocations for each tool usage, as also illustrated in~\fref{fig:teaser}.
In this section, we demonstrate how \OURS can identify parallelizable patterns and execute independent tasks concurrently to resolve this issue.
To do so, we use the following two benchmarks:

\vspace{-4mm}
\begin{itemize}[leftmargin=3mm]
    \item \textbf{HotpotQA:} A dataset that evaluates multi-hop reasoning~\cite{yang2018hotpotqa}. We only use the comparison dev set.
This contains 1.5k questions comparing two different entities, thus exhibiting a 2-way embarrassingly parallel execution pattern. 
An example question is shown in~\fref{fig:teaser}.
\vspace{-4mm}
\item \textbf{Movie Recommendation:} A dataset with 500 examples that asks to identify the most similar movie out of four options to another set of four movies, exhibiting an 8-way embarrassingly parallel pattern~\cite{srivastava2023beyond}.
\end{itemize}

\vspace{-2mm}
\paragraph{Experimental Setups.}
As a baseline method, we compare \OURS with ReAct.
We follow the ReAct setup~\cite{yao2022react} using the same Wikipedia search tool that LLMs can use to search for information.
We did not include the lookup tool since it is not relevant to our problem setting. 
We have optimized the prompt and in-context examples for both ReAct and \OURS to the best of our abilities.
For all experiments across these datasets, we use {gpt-3.5-turbo} (1106 release).
For the experiments using GPT, we additionally report the results using OpenAI's parallel function calling capability, which was announced concurrently with our work.
We also show how \OURS can be effectively combined with the open-source LLaMA-2 70B model to provide the model with parallel function calling capabilities.
For all experiments, we have measured accuracy, end-to-end latency, as well as input and output token usage.
See Appendix~\ref{appendix:details} for details on experimental setups.

\vspace{-2mm}
\paragraph{Accuracy and Latency.}
We report the accuracy, end-to-end latency, and relative speed-up of \OURS compared to ReAct in~\tref{table:benchmark}.
First, we observe that ReAct consistently achieves  
lower accuracy compared to OpenAI parallel function calling and \OURS.
We identify two main failure modes in ReAct: 
(1) the tendency for redundant generation of prior function calls, 
a point also noted in the original ReAct paper~\cite{yao2022react}; 
and (2) premature early stopping based on the incomplete intermediate results.
In Appendix~\ref{subsec:accuracy_analysis}, we offer a detailed analysis demonstrating how these two prevalent failure cases significantly hurt ReAct's accuracy, 
and how they can be resolved with \OURS, leading to an accuracy enhancement of up to 7 -- 8\%.
Furthermore, we have conducted interventional experiments in which we incorporated ReAct-specific prompts to avoid repetitive function calls and early stopping.
ReAct$^\dagger$ in~\tref{table:benchmark} refers to ReAct \textit{with} this ReAct-specific prompt.
The ReAct-specific prompt yields a general accuracy improvement with ReAct$^\dagger$ as compared to the original ReAct.
Nevertheless, \OURS still demonstrates on-par and better accuracy than ReAct$^\dagger$, as such prompting does not serve as a perfect solution to completely avoiding the erroneous behavior of ReAct.

Additionally, when compared to ReAct$^\dagger$, \OURS demonstrates a noticeable speedup of $1.80\times$ and $1.40\times$ on HotpotQA with GPT and LLaMA, respectively.
Similarly, \OURS demonstrates $3.74\times$ and $2.82\times$ speedup on Movie Recommendation with each model. 
Note that we benchmark the latency of \OURS against that of ReAct$^\dagger$ since the repeating and early stopping behavior of the original ReAct as discussed above makes its latency unpredictable and unsuitable for a fair comparison.
\OURS demonstrates a speedup of up to 35\% compared to OpenAI parallel function calling whose latency gain over ReAct is $1.61\times$  and $2.76\times$ on each benchmark.\footnote{
\scriptsize{
Unfortunately, we are unable to conclude why this is the case, as OpenAI has not publicly disclosed any details about their function calling mechanism.
One speculation is that there might be additional overheads to validate the function and argument names and to convert them into a system prompt.
Nevertheless, we have seen a consistent trend with multiple runs over several days. 
}
}

\vspace{-2mm}
\paragraph{Costs.} Another important consideration of using LLMs is cost,
which depends on the input and output token usage.
The costs for GPT experiments are provided in~\tref{table:cost}.
\OURS is more cost-efficient than ReAct for cost, as it involves less frequent LLM invocations. 
Interestingly, \OURS also outperforms the recent OpenAI parallel function calling in cost efficiency.
This is because \OURS's planning phase is more prompt length efficient than that of OpenAI parallel function calling since our Planner's in-context examples are rather short and only include plans, not observations (see Appendix~\ref{appendix:planner-prompt}).

\subsection{Parallel Function Calling with Dependencies}
\label{subsed:multi-tool}
The cases considered above are rather simple, as only one tool is used and all tasks can be executed independently of one another. However, similar to code execution in traditional code blocks, we may encounter function calling scenarios that involve more complex dependencies.
To systematically evaluate the capability to plan out function calling in scenarios that involve complex task dependencies, we have designed a custom benchmark called \BENCH.
This benchmark is designed to incorporate non-trivial function calling patterns, including three different types of patterns in~\fref{fig:parallel-patterns} (b) and (c).
Inspired by the IfQA benchmark~\cite{yu2023ifqa}, \BENCH contains 113 examples that involve
mathematical questions on factual attributes of various entities.
In particular, completing the task requires using two tools
(i.e., search and math tools), with the second tool's argument depending on the result of the first tool's output. 
We have meticulously included questions that are answerable only with information from Wikipedia's first paragraph, effectively factoring out the failure cases due to unsuccessful searches.
See Appendix~\ref{appendix:custom-benchmark} for more details in \BENCH.



\vspace{-2mm}
\paragraph{Experimental Setups.}
Similar to Sec.~\ref{subsec:parallelizable_workloads}, we use ReAct~\cite{yao2022react} as the main baseline.
Here, both ReAct and \OURS are equipped with two tools: 
(1) the {search} tool, identical to the one mentioned in Sec.\ref{subsec:parallelizable_workloads}; and (2) the {math} tool, which solves mathematical problems.
The {math} tool is inspired by the Langchain~\cite{langchain}'s \texttt{LLMMathChain}, which uses an LLM as an agent that interprets input queries and invokes the \texttt{numexpr} function with the appropriate formula.
This enables the {math} chain to address a broad spectrum of math problems that are written both in mathematical and verbal form.
See Appendix~\ref{appendix:details} for more details on experimental setups.

\vspace{-2mm}
\paragraph{Accuracy and Latency.}
As shown in the \BENCH row of~\tref{table:benchmark}, \OURS arrives at
the final answer with an average speedup of $2.15\times$ with gpt-4-turbo and $2.27\times$ with LLaMA-2 70B,
by avoiding sequential execution of the dependency graphs.
Beyond the latency speedup, we observe higher accuracy of \OURS with the LLaMA-2 model as compared to
that of ReAct, due to the reasons discussed in Sec.~\ref{subsec:parallelizable_workloads}.
Particularly in the LLaMA-2 experiment, where \OURS achieves around a 9\% increase in accuracy,
we note that $\sim$20\% of the examples experienced repetitive function calls with ReAct, 
aligning with our observations from the accuracy analysis detailed in Appendix~\ref{subsec:accuracy_analysis}.
Additionally, a comprehensive analysis of \OURS's failure cases is provided in Appendix~\ref{appendix:failure-cases}, where we note minimal Planner failures, highlighting \OURS's effectiveness in breaking down problems into complex multi-task dependencies.

\vspace{-2mm}
\paragraph{Cost.} 
Similar to Sec. \ref{subsec:parallelizable_workloads}, \OURS demonstrates substantial cost reductions of 4.65$\times$ and 2.57$\times$ compared to ReAct and OpenAI's parallel function calling, respectively, as indicated in \tref{table:cost}. 
This efficiency stems from \OURS's reduced frequency of LLM invocations, which is also the case with OpenAI's parallel function calling, which is limited to planning out immediate parallelizable tasks, not the entire dependency graph. 
For example, in \fref{fig:parallel-patterns} (c), OpenAI's method would necessitate three distinct LLM calls for initial search tasks, following math tasks, and the final math task. In contrast, \OURS achieves this with a single LLM call, planning all tasks concurrently.


\vspace{-1mm}
\subsection{Parallel Function Calling with Replanning} 
\label{subsec:replanning-heavy}
\vspace{-1mm}
In the previous sections, we have discussed cases in which dependency graphs can be determined statically.
However, there are cases where dependency graphs need to be constructed dynamically depending on intermediate observations.
Here, we consider one such dynamic approach in the context of the Game of 24 with the Tree-of-Thoughts (ToT) strategy proposed in~\cite{yao2023tree}. 
The Game of 24 is a task to generate 24 using a set of four numbers and basic arithmetic operations. 
For example, from the numbers 2, 4, 4, and 7, a solution could be $4 \times (7 - 4) \times 2 = 24$. 
ToT approaches this task through two iterative LLM processes: (i) the thought proposer generates candidate partial solutions by selecting two numbers and applying an operation (e.g. 2, 3, 7 from 2, 4, 4, 7 by calculating 7 - 4); (ii) the state evaluator assesses the potential of each candidate. 
Only the promising candidates are then processed in subsequent iterations of the thought proposer and state evaluator until 24 is reached.
Details about the Game of 24 benchmark and the ToT strategy can be found in Appendix~\ref{appendix:tot-details}.

While ToT achieves significant improvement at solving the Game of 24, its sequential, breadth-first search approach through the state tree can be time-consuming. 
\OURS offers a faster alternative by enabling parallel execution of the thought proposer and the subsequent feasibility evaluator, akin to a parallel beam search method.

\vspace{-2mm}
\paragraph{Experimental Setups.} 
Although \OURS offers latency advantages, solving this problem with a single static graph is not feasible, 
as the Planner cannot plan out the thought proposing stage before identifying the selected candidates from the state evaluator of the previous iteration.
Consequently, the Planner is limited to planning only within one iteration at a time. 
To address this, we resort to \OURS's replanning capability. 
In particular, \OURS is equipped with three tools: \texttt{thought\_proposer} and \texttt{state\_evaluator}, which are both LLMs adapted from the original ToT framework, 
and \texttt{top\_k\_select}, which chooses the top $k$ candidates from the \texttt{thought\_proposer} based on the \texttt{state\_evaluator}'s assessment. 
After all these tools are executed, \OURS can decide to ``replan'' if no proposal reaches 24, 
triggering the Planner to devise new plans using the shortlisted states from \texttt{top\_k\_select} of the previous iteration.
In this way, \OURS can dynamically regenerate plans of each iteration, being able to tackle highly complex tasks that require iterative replanning based on the outcomes of previous plans.

To evaluate \OURS's performance on the Game of 24, we use 100 different instances of the game.
For each problem, we consider the output as successful if its operations are valid and yield 24 while also using the provided numbers exactly once each.
Further details on experiment setups are outlined in Appendix~\ref{appendix:details}.

\vspace{-2mm}
\paragraph{Success Rate and Latency.} 
In the last two rows of \tref{table:benchmark}, we explore the latency and success rate of \OURS in comparison to the baseline described in~\cite{yao2023tree} on the Game of 24 benchmark. 
With the gpt-4 model, \OURS demonstrates a $2.89\times$ enhancement in latency while slightly improving the success rate compared to the baseline. Similarly, when applied with the LLaMA-2 model, \OURS shows a $2.01\times$ improvement in latency, again without compromising on success rate. 
These results demonstrate not only a significant latency reduction without quality degradation, but also the replanning capability of \OURS for solving complex problems.

\subsection{Application: \OURS in Interactive Decision Making Tasks} 
\label{subsec:interactive_decision_making_tasks}
In this section, we demonstrate that \OURS can explore language-based interactive environments effectively by benchmarking \OURS on WebShop~\cite{yao2022webshop}.
As highlighted in \cite{shinn2023reflexion, yao2022react, yao2022webshop}, WebShop exhibits considerable diversity, which requires extensive exploration to purchase the most appropriate item.
While recent work feature advanced exploration strategies and show promising results~\cite{zhou2023language,ma2023laser}, their approaches are largely based on a sequential and extensive tree search that incurs significant latency penalties. 
Here, \OURS showcases an exploration strategy that is both effective and efficient with the use of parallel function calling. 
Our method enables broader exploration of items in the environment, which improves success rate compared to ReAct.
At the same time, this exploration can be parallelized, yielding up to $101.7\times$ speedup against baselines that perform sequential exploration.

\vspace{-2mm}
\paragraph{Experimental Setups.} 
We evaluate \OURS against three baselines on this benchmark, ReAct~\cite{yao2022react}, LATS~\cite{zhou2023language}, and LASER~\cite{ma2023laser}, using 500 WebShop instructions. 
The evaluation metrics are success rate, average score, and latency.
More details of the WebShop environment and the baseline methods are provided in Appendix~\ref{appendix:webshop-details}.
For this experiment, \OURS is equipped with two tools: \texttt{search} and \texttt{explore}. The \texttt{search} function triggers the model to generate and dispatch a query that returns a list of typically ten items from the Webshop environment. 
The \texttt{explore} function then clicks through links for each of the found items and retrieves information about options, prices, attributes, and features that are available. Finally, based on the gathered information, \OURS decides on the item that best matches the input instruction for purchasing. Further details on experiments can be found in Appendix~\ref{appendix:details}.

\vspace{-2mm}
\paragraph{Performance and Latency.} 

Our approach significantly outperforms all baseline models as shown in Table~\ref{table:webshop}.
When using gpt-3.5-turbo, \OURS achieves a 28.4$\%$ and 6$\%$ improvement in success rate against ReAct and LATS;
with gpt-4, our method improves upon ReAct and LASER by 20.4$\%$ and 5.6$\%$, respectively.
In terms of latency, \OURS exhibits a $101.7\times$ and $2.69\times$  speedup against LATS and LASER.
While we note that \OURS execution is slightly slower than ReAct on this benchmark, mainly due to the Planner overhead,
we also highlight that the gains in success rate far outweigh the minor latency penalty.

We further delve into why \OURS attains such an improved success rate and score compared to ReAct.
Based on our observations, we discover that the ReAct agent tends to commit to a decision with imperfect information, a scenario that can arise when the agent has not gathered sufficient details about the features and options available for items. 
This observation was also noted in \cite{shinn2023reflexion} -- without exploring more items in the environment, the agent struggles to differentiate between seemingly similar choices, ultimately failing to make the correct decision.
In contrast, \OURS undergoes further exploration by visiting all ten items found by \texttt{search} and retrieving relevant information about each item. 
We find that employing an effective search strategy is critical to decision-making tasks such as the WebShop benchmark.

\begin{table}[!t] 
\vspace{-2mm}
\caption{Performance and Latency Analysis for WebShop. We evaluate \OURS  with two models: gpt-4 and gpt-3.5-turbo and compare \OURS against three baselines: ReAct, LATS, and LASER. We report success rate and average score in percentage. We reproduce the success rate and average score for ReAct, while those for LATS and LASER are from their papers. N denotes the number of examples used for evaluation.
}

\vspace{-1mm}
\begin{center}
\scriptsize{
\setlength{\tabcolsep}{5pt}{
\begin{tabular}{c|c|cc|c|c}
\toprule

 {Model} &  {Method} &  {Succ. Rate} & Score & Latency (s)  & {N}   \\
\midrule
 \multirow{4}{*}{gpt-3.5-turbo} & ReAct & 19.8  & 54.2 & 5.98 & 500 \\ 
   & LATS & 38.0  & 75.9 & 1066 & 50 \\ 
   & \graycell{\OURS} &  \graycell{44.0} & \graycell{72.8} & \graycell{10.72} & \graycell{50}\\
   & \graycell{\OURS} &  \graycell{48.2} & \graycell{74.2} & \graycell{10.48} & \graycell{500}\\
\midrule
\multirow{3}{*}{gpt-4-0613} & ReAct & 35.2 & 58.8 & 19.90 & 500 \\ 
   & LASER & 50.0  & 75.6 & 72.16 & 500\\ 
   & \graycell{\OURS} &  \graycell{55.6} & \graycell{77.1} & \graycell{26.73} & \graycell{500}\\
 
\bottomrule
\end{tabular}
}
}
\end{center}
\vspace{-6mm}
\label{table:webshop}
\end{table}

The relatively high performance of LATS can also be explained in terms of its exploration scheme.
In this framework, the agent executes a brute-force search through the state and action space of Webshop, exploring as many as 30 trajectories before making the final purchase.
While this approach provides richer information for decision-making, the end-to-end execution becomes prohibitively slow.

We report that our method, \OURS, outperforms LASER by an average score of 1.5. When compared to LATS, this score is within the standard deviation range of our method. The average score for \OURS, along with its standard deviation, is $72.8\pm4.01$ for gpt-3.5-turbo. 
Further note that while the performance differences are marginal, our method exhibits significant execution speedup, $101.7\times$ over LATS and $2.69\times$ over LASER.

\section{Conclusions}
Existing methods for invoking multiple functions with LLMs resort to sequential and dynamic reasoning. As a result, 
they suffer from inefficiencies in latency, cost, and accuracy.
As a solution, we introduced \OURS, a compiler-inspired framework that enables efficient parallel function calling across various LLMs, including open-source models like LLaMA-2 and OpenAI's GPT series. 
By decomposing user inputs into tasks with defined inter-dependencies and executing these tasks concurrently through its Planner, Task Fetching Unit, and Executor components, 
\OURS demonstrates substantial improvements in latency (up to $3.7\times$), cost efficiency (up to $6.7\times$), and accuracy (up to ${\sim}9\%$), even outperforming OpenAI's parallel function calling feature in latency gains. 
We look forward to future work building upon \OURS that will improve both the capabilities and efficiencies of LLMs in executing complex, large-scale tasks, thus transforming the future development of LLM-based applications.

\subsection*{Impact Statement}
This paper presents research towards advancing the field of Machine Learning.
While there are many potential societal consequences of our work, we do not find any one to be particularly noteworthy.

\subsection*{Acknowledgements}
We appreciate the valuable feedback from Minwoo Kang.
We acknowledge gracious support from Furiosa team.
We also appreciate the support from Microsoft through their Accelerating Foundation Model Research, including
great support from Sean Kuno.
Furthermore, we appreciate support from
Google Cloud, the Google TRC team, and specifically Jonathan Caton, and Prof. David Patterson.
Prof. Keutzer's lab is sponsored by the Intel corporation, Intel One-API, Intel VLAB team, the Intel One-API center of
excellence, as well as funding through BDD and BAIR.
We also appreciate support from Samsung including Dongkyun Kim, and David Thorsley.
We appreciate the great support from Ellick Chan, Saurabh Tangri, Andres
Rodriguez, and Kittur Ganesh.
Sehoon Kim and Suhong Moon would like to acknowledge the support from the Korea Foundation for Advanced Studies.
Amir Gholami was supported through funding from Samsung SAIT.
Michael W. Mahoney would also like to acknowledge
a J. P. Morgan Chase Faculty Research Award 
as well as 
the DOE, NSF, and IARPA.
Our conclusions do not necessarily reflect the position or the policy of our sponsors, and no official endorsement should be~inferred.

\bibliography{paper}
\bibliographystyle{icml2024}

\renewcommand\thefigure{\thesection.\arabic{figure}} 
\setcounter{figure}{0} 

\renewcommand\thetable{\thesection.\arabic{table}} 
\setcounter{table}{0} 
\newpage
\appendix
\clearpage
\onecolumn

\section{Accuracy Analysis: ReAct vs. \OURS}
\label{subsec:accuracy_analysis}
In this section, we delve into a detailed analysis that compares the accuracy of both ReAct and \OURS, highlighting two failure cases that are prevalent in ReAct: (i) premature early stopping; and (ii) repetitive function calls.
Furthermore, we demonstrate that while those failure cases negatively impact the ReAct accuracy, they can be effectively addressed by \OURS, thereby yielding the improved accuracy of our framework. 
We analyze two specific scenarios: the Movie Recommendation evaluation with GPT, where ReAct often prematurely stops, leading to significantly lower accuracy compared to \OURS (68.60 vs. 77.13 in~\tref{table:benchmark}); and the HotpotQA evaluation with LLaMA-2 70B, where ReAct's repetitive function calls result in a notable accuracy degradation compared to \OURS (70.00 vs. 77.80 in~\tref{table:benchmark}).


\begin{figure*}[h]
\vspace{-2mm}
\centering
\includegraphics[width=0.8\linewidth]{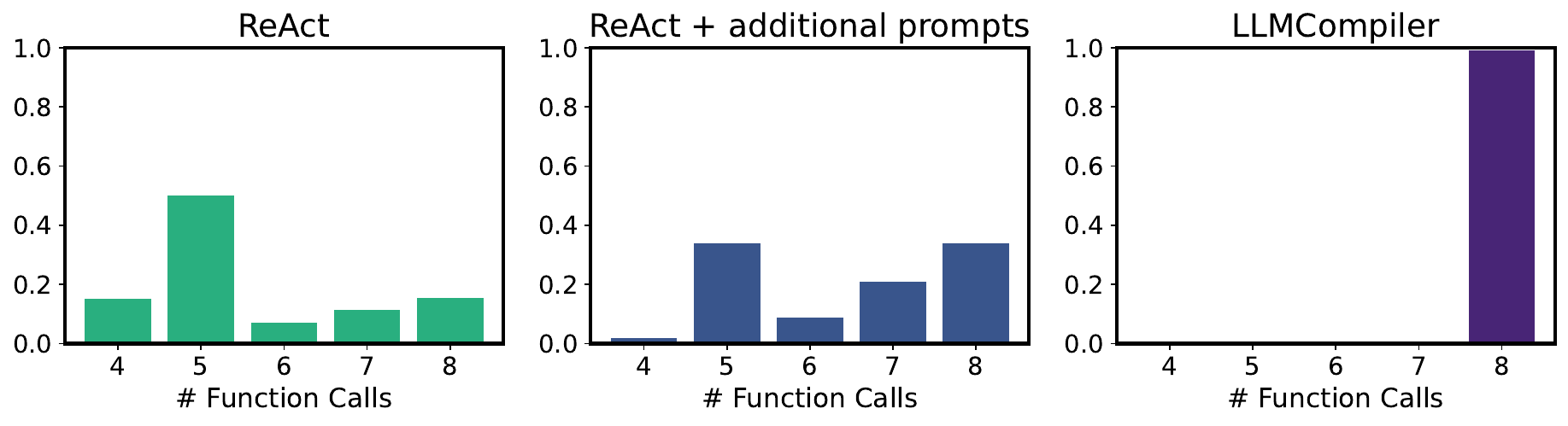}
\caption{Distributions of the number of function calls when running the Movie Recommendation benchmark on ReAct (Left), ReAct with specific prompts to avoid early stopping (Middle, corresponding to ReAct$^\dagger$ in~\tref{table:benchmark}), and \OURS (Right). 
\OURS (Right) consistently completes the search for all 8 movies, whereas ReAct (Left) often exit early, demonstrated by about 85\% of examples stopping early.
Although the custom prompts shift ReAct's histogram to higher function calls (Middle), they still fall short of ensuring comprehensive searches for all movies.
gpt-3.5-turbo is used for the experiment.
  }
  \label{fig:movie-dist}
\end{figure*}



\begin{figure*}[h]
\vspace{-2mm}
\centering
\includegraphics[width=0.35\linewidth]{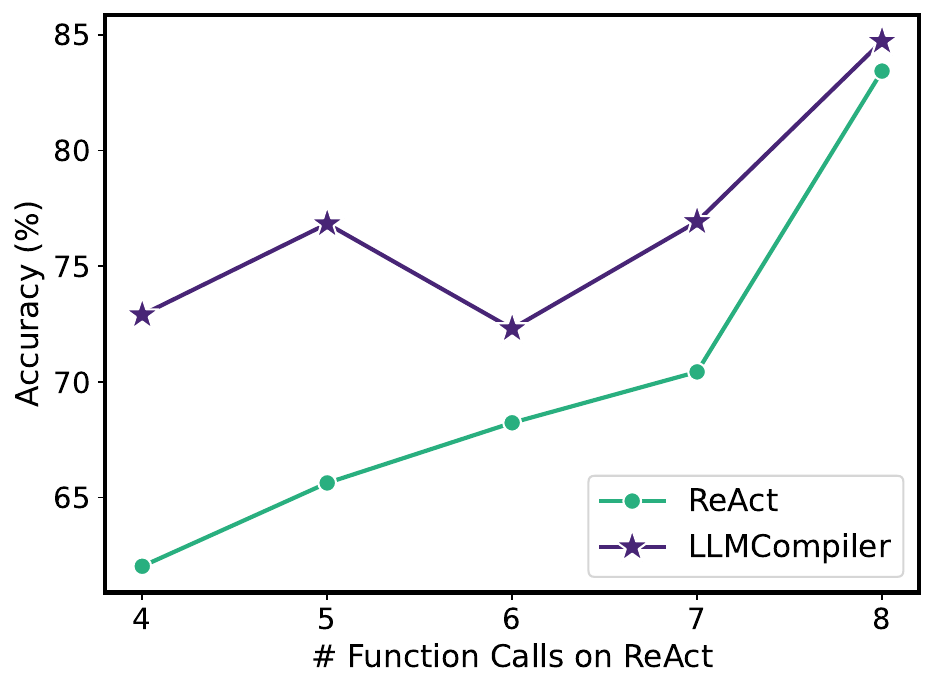}
\caption{
The Movie Recommendation accuracy of the examples that are categorized by the number of function calls on ReAct, measured both on ReAct and \OURS.
The plot indicates that in ReAct,  a decrease in the number of function calls correlates with lower accuracy, indicating that premature exits lead to reduced accuracy.
In contrast, when the same examples are evaluated using \OURS, which ensures complete searches for all eight movies before reaching a decision, they achieve higher and more consistant accuracy than those processed by ReAct.
gpt-3.5-turbo is used for the experiment, and the results are averaged over 3 different runs.
  }
 \vspace{-3mm}
  \label{fig:movie-acc}
\end{figure*}


\paragraph{Premature Early Stopping of ReAct. }
ReAct frequently suffers from premature early stopping, ceasing function calls too early, and making decisions based on incomplete information. 
A clear example of this is observed in the Movie Recommendation benchmark, where ReAct often searches for fewer than the required 8 movies before delivering its final answer.
 In~\fref{fig:movie-dist} (Left), we illustrate the distribution of the number of function calls within ReAct (using GPT) across thhe Movie Recommendation benchmark.
 Here, we observe around 85\% of the examples exhibit early stopping, making decisions without completing all 8 movie searches. 
 This contrasts with \OURS (Right), where almost all examples (99\%) complete the full search of 8 movies. 
 Although adding specific prompts to ReAct to prevent early stopping shifts the distribution towards more function calls (\fref{fig:movie-dist}, Middle), resulting in an accuracy improvement from 68.60 to 72.47 (ReAct$^\dagger$ in~\tref{table:benchmark}), it is nevertheless an imperfect solution.

To further assess how early stopping negatively impacts accuracy, we categorize Movie Recommendation benchmark examples by their number of function calls in ReAct. 
We then evaluated these groups using \OURS, ensuring complete search results for all 8 movies. 
\fref{fig:movie-acc} reveals that fewer function calls in ReAct correlate with lower average accuracy (green line). 
Conversely, if these examples were processed through \OURS, with complete searches for all eight movies, they consistently attained higher accuracy (purple line). 
This not only indicates that ReAct struggles with premature exits (which is not fully addressed by prompting), but the earlier it stops, the greater the decline in accuracy, contributing to the overall accuracy drop observed in~\tref{table:benchmark}. 
In contrast, \OURS effectively addresses this issue.


\begin{figure*}[!t]
\vspace{-2mm}
\centering
\includegraphics[width=0.6\linewidth]{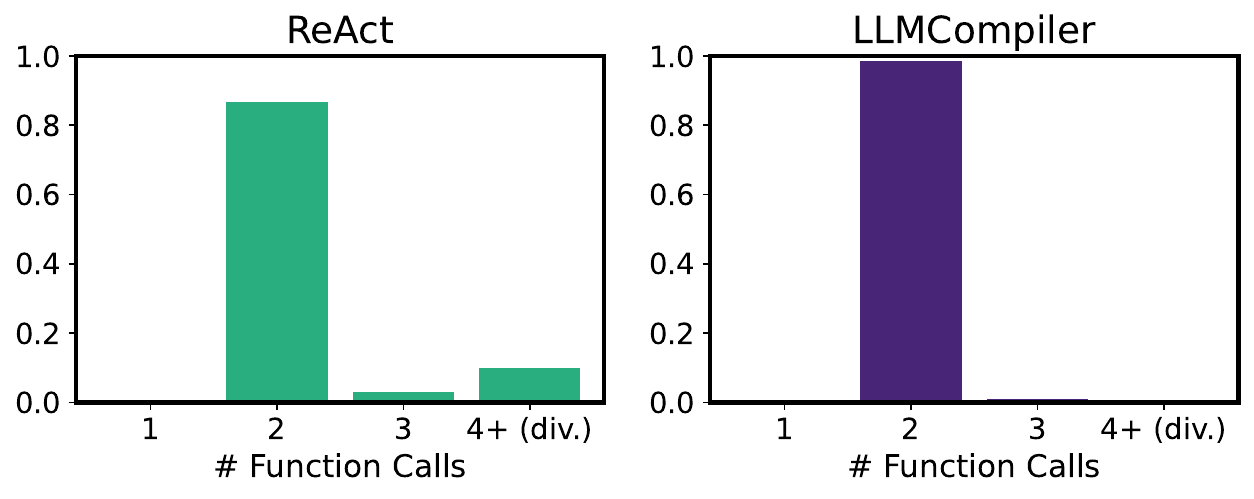}
\caption{Distributions of the number of function calls when running the HotpotQA benchmark on ReAct (Left) and \OURS (Right). 
While \OURS (Right) consistently completes the task within 2 function calls, which is expected as HotpotQA exhibits a 2-way parallelizable pattern, 
ReAct (Left) shows that around 10\% of the examples undergo repetitive ($>$4) function calls, resulting in a diverging behavior of the framework.
LLaMA-2 70B is used for the experiment.
  }
  \label{fig:hotpot-dist}
\end{figure*}



\begin{figure*}[!t]
\vspace{-2mm}
\centering
\includegraphics[width=0.35\linewidth]{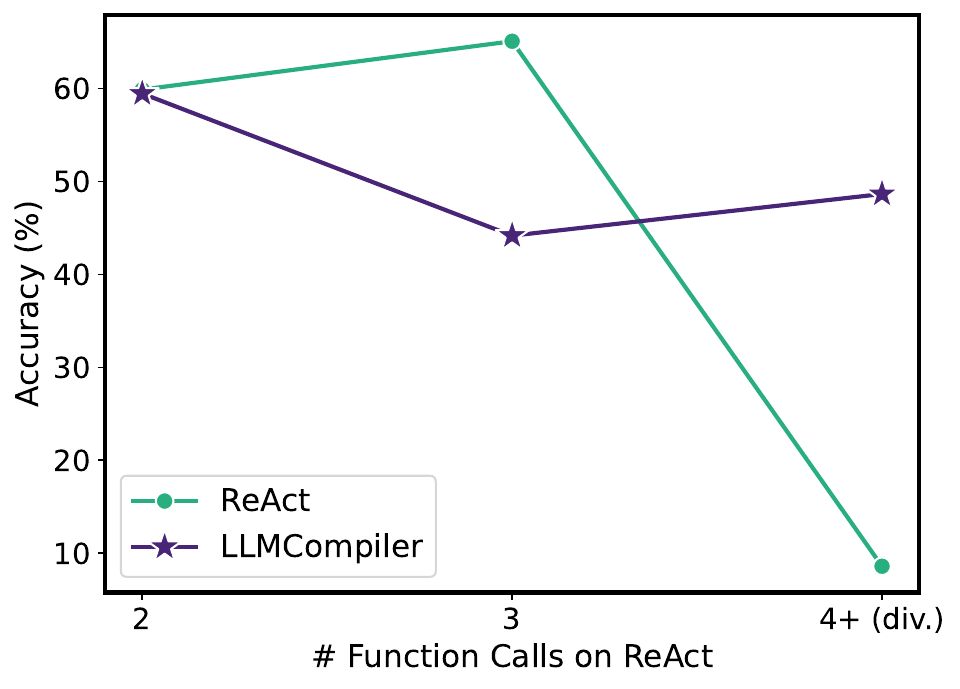}
\caption{
The HotpotQA accuracy of the examples that are categorized by the number of function calls on ReAct, measured both on ReAct and \OURS.
The plot indicates that in ReAct, repetitive function calls of more than or equal to four times can result in a significant accuracy degradation due to its infinite looping and diverging behavior.
On the other hand, when the same examples are evaluated using \OURS, which ensures only two searches per example, they achieve a higher of around 50\%.
LLaMA-2 70B is used for the experiment.
  }
 \vspace{-3mm}
  \label{fig:hotpot-acc}
\end{figure*}


\paragraph{Repetitive Function Calls of ReAct. }
Another common failure case of ReAct is its tendency for repetitive function calls, often leading to infinite loops or exceeding the context length limit. 
 This problem is particularly noticeable in the HotpotQA benchmark where ReAct repeatedly calls the same function if the Wikipedia search returns insufficient information about the searched entity. 
Although HotpotQA is inherently 2-way parallelizable, as illustrated in~\fref{fig:hotpot-dist}, we observe that about 10\% of its examples require more than four function calls in ReAct, usually resulting in an infinite loop or a divergent behavior. 
In contrast, \OURS executes only two function calls for most examples.

To show how the repetitive function calls impact the overall accuracy, we conduct an accuracy analysis similar to the previous case.
In~\fref{fig:hotpot-acc}, we categorize HotpotQA benchmark examples by the number of function calls in ReAct, and then we compare their accuracy on both ReAct and \OURS. 
The analysis reveals that examples that launch two function calls in ReAct maintain the same accuracy in \OURS.
However, cases with more than four function calls in ReAct, which often lead to divergent behavior, show less than 10\% accuracy in ReAct.
On the other hand, when these examples are processed with \OURS, they achieve around 50\% accuracy by circumventing repetitive calls. 
It is worth noting that there are instances with three function calls in ReAct, where an extra search can lead to improved accuracy by retrying with an alternate entity name when the initial search fails, yielding a better accuracy than \OURS.
While this shows a potential adaptability advantage of ReAct, such instances represent less than 3\% of cases.

\section{Failure Case Analysis of \OURS}
\label{appendix:failure-cases}
This section delves into a qualitative analysis of \OURS's failure cases on the \BENCH benchmark, which can be broadly attributed to failures in the Planner, Executor, or the final output process. 
Failures in the final output process refer to cases when LLMs are unable to use the observations collected from tool execution (which are incorporated into the context) to deliver the correct answer to the user. 
Among the 10.6\% (36 examples) of \OURS's total failures reported in Tab.~\ref{table:benchmark}, we have noted that the Planner, Executor, and final output process contributed to 8\%, 64\%, and 28\% of the failures, respectively. 
The Planner's 8\% failure rate is exclusive to \OURS.
For instance, the Planner would incorrectly map inputs and outputs by assigning a wrong identifier as an input to a subsequent task, thereby forming an incorrect DAG.
However, with adequate tool definitions and in-context examples, Planner errors are significantly reduced (only 3 instances in total throughout our evaluation), highlighting the LLM's capability to decompose problems into complex multi-task dependencies.

The remaining 92\% of the total failures are attributed to the Executor and the final output process. 
The Executor accounts for most of these failures (64\%), with common issues like the \texttt{math} tool choosing wrong attributes or mishandling unit conversions.
For the final output process (28\% of failures), errors include incorrect conclusions from the gathered observations, such as failing to pick the smallest attribute from the collected data. 
It's worth noting that these problems are not exclusive to \OURS, but they also occur in ReAct.
Nevertheless, \OURS tends to have slightly fewer failures in these areas than ReAct, as it provides only relevant contexts to each tool, aiding in more accurate information extraction.
We believe that optimizing the structure of the agent scratchpad, rather than simply appending observations, could further reduce failures in the final output process.

\section{Related Work}
Here, we continue with related work, which we started in Sec.~\ref{sxn:related_work_x}.
\subsection{Tool-Augmented LLMs}
\label{appendix:tool-augmented-llms}

A notable work is Toolformer~\cite{schick2023toolformer}, which produces a custom LLM output to let the LLM decide what the inputs for calling the functions should be and where to insert the result.
This approach has inspired various tool calling frameworks~\cite{liang2023taskmatrixai, shen2023hugginggpt}. 
ReAct~\cite{yao2022react} proposed to have LLMs interact with external environments through reasoning and action generation for improved performance.
Gorilla \cite{patil2023gorilla} introduced a finetuned LLM designed for function calling, and ToolLLM~\cite{qin2023toolllm} and RestGPT \cite{song2023restgpt} have extended LLMs to support real-world APIs.
Moreover, OpenAI \cite{openai2023gpt4} released their own function calling capabilities, allowing their LLMs to return formatted JSON for execution.

\begin{table*}[!t]
\caption{
A latency comparison between using and not using streaming in the Planner. 
Streaming yields consistent latency improvement across different benchmarks, as it enables the Task Fetching Unit to start task execution immediately as each task is produced by the Planner. The impact of streaming is especially notable in the \BENCH benchmark, where tool execution times are long enough to effectively hide the Planner's execution time.
}
\vspace{-1mm}
\begin{center}
\footnotesize{
\setlength{\tabcolsep}{6.5pt}{
\begin{tabular}{c|cc|c}
\toprule

 Benchmark & w/o streaming (s) &  w/ streaming (s) & Latency speedup \\
\midrule
HotpotQA & 4.00 & \textbf{3.95} & 1.01$\times$ \\
Movie Rec. & 5.64 & \textbf{5.4}7 & 1.03$\times$ \\
\BENCH & 21.72 & \textbf{16.69} & 1.30$\times$ \\
\bottomrule
\end{tabular}
}
}
\end{center}
\vspace{-4mm}
\label{table:streaming}
\end{table*}


\section{Experimental Details}
\label{appendix:details}

Our experiments evaluate two different common scenarios: (1) using API-based closed-source models; and (2) using open-source models with an in-house serving framework. We use OpenAI's GPT models as closed-source models, in particular, gpt-3.5-turbo (1106 release) for HotpotQA and Movie Recommendation, gpt-4-turbo (1106 release) for \BENCH, and gpt-4 (0613 release) for Game of 24. Experiments on HotpotQA, Movie Recommendation, and \BENCH were all conducted in November 2023 after the 1106 release. The Game of 24 experiments were conducted over a two-month period from September to October 2023.
For an open-source model, we use LLaMA-2~\cite{touvron2023llama}, which was hosted on 2 A100-80GB GPUs using the vLLM~\cite{kwon2023vllm} framework.
All the runs have been carried out with zero temperature, except for \texttt{thought\_proposer} and \texttt{state\_evaluator} for the Game of 24 evaluation, where the temperature is set to 0.7.
Since OpenAI has randomness in outputs even with temperature 0, we have conducted 3 runs, and we reported the average accuracy.
Across ReAct, OpenAI parallel function calling, and \OURS, we perform 3, 1, and 5-shot learning for HotpotQA, Movie Recommendation, and \BENCH, respectively; the same examples across different methods were used to ensure a fair comparison.
For the Game of 24, we use 2 in-context examples for the Planner.
We use the same instruction prompts across different methods for a fair comparison, except for ReAct$^\dagger$ in Sec.~\ref{subsec:parallelizable_workloads} with additional ReAct-specific prompts. For WebShop experiment, we use gpt-4-0613 with 8k context window and gpt-3.5-turbo model with 16k context window. 

\section{Analysis}
\label{sec:discussions}

\subsection{Parallel Speedup Modeling}
\label{sec:latency-modeling}
    
While \OURS shows noticeable latency gain in various workloads, it is not achieving the $N\times$ latency speedup for N-way parallel workloads. 
This is mostly due to the overhead associated with \OURS's Planner and final answering process that cannot be parallelized. 
In our Movie Recommendation experiment, \OURS's Planner and the answering process have an overhead of 1.88 and 1.62 seconds on average, respectively, whose combined overhead already comprises more than half of \OURS's overall latency in Tab~\ref{table:benchmark}. 
Another source of overhead is the straggler effect among the parallel tasks when they need to join together.
We observe the average latency of the slowest \texttt{search} to be 1.13 seconds, which is nearly $2 \times$ the average latency of all tasks, which is 0.61 seconds. 
Below, we provide an analytical latency modeling of ReAct, \OURS, and \OURS with streaming, and we provide an analysis of achievable latency speedup.

In this section, our focus is on \emph{embarrassingly parallelizable} workload (pattern~\fref{fig:parallel-patterns}(a)), 
as this allows for a clearer understanding of the impact 
of each component on potential latency gains. 
For the precise latency analysis, 
we consider three key components: the Planner, the Task Fetching Unit, and the Executor, in~\fref{fig:system_overview}.
Assume that the Planner generates $N$ different tasks to be done. 
We define $P_i$ as the Planner's output corresponding to the $i$-th atomic task. Each $P_i$ is a blueprint for a specific atomic task, which we refer to as $E_i$. 
The execution of $E_i$ involves a specific function call using the appropriate tool.
The latency function of each unit in the system is defined to quantify the time taken for specific operations.
For the Planner, the latency is denoted as $T_P(P_i)$, representing the time taken by the Planner to generate the plan $P_i$. 
Similarly, for the Executor, the latency, $T_E(E_i)$, corresponds to the time required to complete the task $E_i$. 
We ignore the latency of Task Formulation Unit, as it is negligible in this section.
Our focus here is on comparing the latency models of ReAct~\cite{yao2022react}, and \OURS. 

To begin our analysis of ReAct's latency, we express its total latency as:
\begin{align}
    T^R = \sum_{i=1}^N \left(T_P^R(P_i) + T_E(E_i)\right) .
    \label{eq:react_latency_model}
\end{align}
Here, the superscript $R$ refers to ReAct. In the ReAct agent system, the process typically involves initial thought generation, 
followed by action generation and the acquisition of observations through function calls associated with the tool. 
The creation of both thought and action are collectively considered as part of generating $P_i$. 
It is important to note that while the Planner's latency is denoted with a superscript (indicating ReAct), 
the Executor's latency does not have such a superscript. 
This is because the function calling and the tools execution remain the same between ReAct and \OURS.

For \OURS, where all parallelizable tasks are processed concurrently, the total latency is determined by the slowest task among these tasks. Hence, the latency model for \OURS can be represented~as:
\begin{align}
T^C = \sum_{i=1}^N T_P^C(P_i) + \max_{k \in {1,\dots,N}} T_E(E_k) .
\end{align}
This expression captures the sum of all planning times plus the execution time of the longest task, reflecting the system's focus on parallel execution.

Further, if the Planner employs streaming of the dependency graph, the latency model undergoes a modification and can be expressed as:
\begin{align}
T^{SC}= \sum_{i=1}^N T_P^C(P_i) + T_E(E_N) .
\end{align}
It is important to note that $T^{SC} \leq T^C$.
This implies that the streaming mechanism allows for a more efficient handling of task dependencies, potentially reducing overall latency. 

In evaluating the potential speedup achievable with the \OURS framework compared to ReAct, the speedup metric, denoted as $\gamma$, is defined as follows:
\begin{align}
\gamma
= \frac{T^R}{T^C}
= \frac{\sum_{i=1}^N \left(T_P^R(P_i) + T_E(E_i)\right)}{\sum_{i=1}^N T_P^C(P_i) + \max_{k \in {1,\dots,N}} T_E(E_k)} .
\label{eq:speedup}
\end{align}
This ratio represents the comparative efficiency of \OURS over ReAct, considering both planning and execution latencies.

To estimate the upper bound of this speedup, $\gamma_{\text{max}}$, we assume that the executor latency $T_E(E_i)$ is dominant over the planning latency $T_P(P_i)$ and all the latencies of executing tasks remain the same.
Under this assumption, the upper bound is calculated as:
\begin{align}
\gamma_\text{max}
\approx \frac{\sum_{i=1}^N T_E(E_i)}{\max_{k \in {1, \dots, N}}T_E(E_k)}
= N ,
\label{eq:upper_bound}
\end{align}
indicating the theoretical maximum speedup, $\gamma_{\text{max}}$, is equal to the number of tasks, $N$.

On the other hand, the lower bound of the speedup, $\gamma$, is observed when the planning latency is the predominant factor. Given that the planning latencies of both ReAct and \OURS are generally similar, the minimum speedup is approximated~as:
\begin{align}
\gamma_\text{min}
\approx \frac{\sum_{i=1}^N T_P^R(P_i)}{\sum_{i=1}^N T_P^C(P_i)} \approx 1.
\label{eq:lower_bound}
\end{align}

From these observations, we can conclude that to achieve significant latency gains with \OURS, it is crucial to (i) reduce the planner overhead and (ii) minimize the occurrence of stragglers.

\subsection{Latency versus Number of Parallelizable Tasks}
\label{subsec:latency-and-parallel-tasks}


\begin{figure}[t!]
\centering
\includegraphics[width=0.35\linewidth]{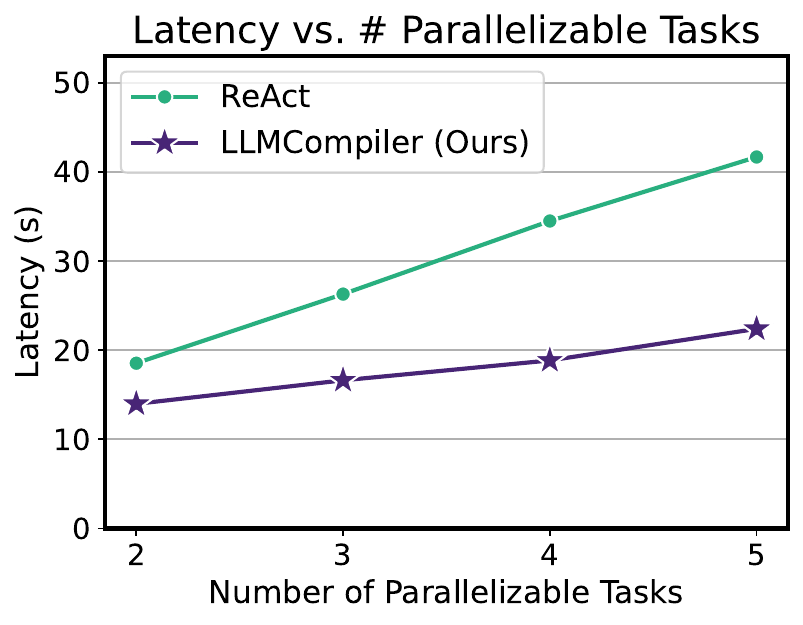}
\vspace*{-5mm}
\caption{Latency on the \BENCH benchmark grouped by the number of maximum parallelizable tasks.
  }
 \vspace{-3mm}
  \label{fig:speedup-breakdown}
\end{figure}

In~\fref{fig:speedup-breakdown}, we also report a more detailed latency breakdown on ParallelQA where 
we show the end-to-end latency as a function of the number of parallel tasks.
This is often referred to as weak-scaling in high-performance computing,
where the ideal behavior is to have a constant latency as the number of 
tasks is increased. We can see that ReAct's latency increases
proportionally to the number of tasks, which is expected as it
executes the tasks sequentially. In contrast, the latency of \OURS
increases at a much smaller rate, as it can perform multiple function calls in parallel
when possible.
The reason the end-to-end latency increases slightly with
\OURS is due to the overhead of the Planner, which needs to generate plans initially, and which cannot be parallelized. 
We provide a further analysis of this in Appendix~\ref{sec:latency-modeling}.


\begin{table*}[!t]
\caption{
Accuracy and latency comparison of \OURS compared to ReAct on the HotpotQA bridge benchmark.
ReAct$^\dagger$ denotes ReAct with additional prompting that minimizes looping and early stopping, similar to~\tref{table:benchmark}.
}
\vspace{-1mm}
\begin{center}
\footnotesize{
\setlength{\tabcolsep}{6.5pt}{
\begin{tabular}{c|cc}
\toprule

 Method & Accuracy (\%) &  Latency (s) \\
\midrule
ReAct & 22.7 & 7.07 \\
ReAct$^\dagger$ & 23.1 & 6.42 \\
\hc \OURS & \textbf{26.3} & \textbf{4.70} \\
\bottomrule
\end{tabular}
}
}
\end{center}
\vspace{-4mm}
\label{table:bridge}
\end{table*}


\subsection{Additional Experiments on the HotpotQA Bridge Benchmark}

In our main experiments in \sref{subsec:parallelizable_workloads}, we used the comparison benchmark in HotpotQA to demonstrate the capability of \OURS in efficiently executing 2-way parallelizable workloads. 
The other part of the benchmark, called `bridge,' involves sequential tasks such as ``What government position was held by the woman who portrayed Corliss Archer in the film Kiss and Tell?" 
\OURS is not limited to the comparison benchmark, but it can also be applied to the bridge benchmark due to its replanning capability: initially, it searches for the woman who played Corliss Archer in the film Kiss and Tell, and then, through replanning, searches the government position held by this woman for the example above.

Similar to our experiments with the comparison benchmark, \tref{table:bridge} compares \OURS against ReAct and ReAct with the additional prompt that avoids repetitive function calling and early stopping (ReAct$^\dagger$) on the bridge benchmark.
We observe 4 and 3\% accuracy improvement, respectively, which is attributed to ReAct’s repetitive function invocation -- even with the additional prompt (ReAct$^\dagger$), we have still observed 5\% of the examples failing with this issue.
Furthermore, such repetitive function call also accounts for the slightly higher latency of ReAct compared to ours.
This experiment demonstrates that \OURS allows for efficient and accurate function calling for both parallel and sequential workloads.


\begin{table*}[!t]
\caption{
Qualitative comparison between \OURS and other frameworks including ReAct~\cite{yao2022react}, TPTU (SA for Sequential Agent and OA for One-step Agent)~\cite{ruan2023tptu}, ViperGPT~\cite{suris2023vipergpt} and HuggingGPT~\cite{shen2023hugginggpt}. 
}
\vspace{-1mm}
\begin{center}
\footnotesize{
\setlength{\tabcolsep}{6.5pt}{
\begin{tabular}{c|cccc}
\toprule

 Method & Planning & Replanning & Parallel Execution & Domain \\
\midrule
ReAct & X & - & X & All \\
TPTU-SA & X & - & X & All \\
TPTU-OA & O & X & X & All \\
ViperGPT & O & X & X & Limited \\
HuggingGPT & O & X & O & Limited \\
\hc \OURS & O & O & O & All\\
\bottomrule
\end{tabular}
}
}
\end{center}
\vspace{-4mm}
\label{table:comp-summary}
\end{table*}


\section{Additional Discussions about Related Works}
\label{appendix:baseline-comparison}

TPTU~\cite{ruan2023tptu}, HuggingGPT~\cite{shen2023hugginggpt}, and ViperGPT~\cite{suris2023vipergpt} have introduced end-to-end plan-and-solve frameworks. 
In this section, we discuss how \OURS distinguishes itself from other frameworks from various angles, including the capabilities in (i) planning and replanning; (ii) parallel execution; and (iii) addressing a wider range of problem domains.
Refer to~\tref{table:comp-summary} for the summary.

\vspace{-1mm}
\noindent
\textbf{Parallel Execution:}
Parallel execution is a critical feature in the \OURS framework that allows for efficient function calling and job completion.
While the One-step Agent in TPTU (i.e., TPTU-OA) incorporates planning, it does not enable parallel function calling, as it only decomposes a user input into a sequence of functions and the associated arguments without their inter-dependencies. 
ViperGPT generates Python codes. However, ViperGPT, by itself, does not support parallel execution without a dedicated parallel processing engine since the standard Python interpreter lacks support for parallel execution. 
While HuggingGPT enables parallel execution, it strictly targets models in HuggingFace, making it hard to apply in a wide range of problems and domains that LLMCompiler supports.

\vspace{-1mm}
\noindent
\textbf{Planning and Replanning:}
The TPTU's Sequential Agent (i.e., TPTU-SA) is an iterative framework like ReAct~\cite{yao2022react}  that executes one action per iteration.
While TPTU-OA, HuggingGPT, and ViperGPT are all planning-based frameworks that plan out multiple actions prior to execution, they lack replanning capabilities.
\OURS, in contrast, incorporates the replanning mechanism to generate a new set of tasks when the previous plans are not sufficient enough to deliver the response back to the user.
This enables \OURS to adapt plans based on intermediate results that are a priori unknown, without the need for introducing complex branching logic, thereby extending the scope of problems that it can address.

\vspace{-1mm}
\noindent
\textbf{Problem Domains:}
ViperGPT and HuggingGPT aim for vision tasks via Python code generation and models in HuggingFace, respectively,
showing significant promise in these specific areas. 
In contrast, \OURS targets a general framework that enables efficient and accurate function calling in a wide range of problem domains, rather than restricting itself to specific fields.


\begin{table*}[!t]
\caption{
Accuracy and latency speedup comparison of \OURS compared to ReAct and TPTU (SA for
Sequential Agent and OA for One-step Agent) on the HotpotQA comparison benchmark using gpt-3.5-turbo.
ReAct$^\dagger$ and TPTU-SA$^\dagger$ denote ReAct and TPTU-SA with additional prompting that minimizes looping and early stopping, respectively, similar to~\tref{table:benchmark}.
}
\vspace{-1mm}
\begin{center}
\footnotesize{
\setlength{\tabcolsep}{6.5pt}{
\begin{tabular}{c|cc}
\toprule

 Method & Accuracy (\%) &  Speedup \\
\midrule
ReAct & 61.52 & - \\
ReAct$^\dagger$ & 62.47 & 1$\times$\\
TPTU-SA & 34.16 & - \\
TPTU-SA$^\dagger$ & 44.59 & 1.09$\times$\\
TPTU-OA & 57.50 & 1.35$\times$\\
\hc \OURS & \textbf{62.00} & \textbf{1.51$\times$} \\
\bottomrule
\end{tabular}
}
}
\end{center}
\vspace{-4mm}
\label{table:tptu}
\end{table*}


\subsection{Quantitative Comparison between \OURS and TPTU}

Additionally, in~\tref{table:tptu}, we additionally provide accuracy and latency speedup of \OURS against TPTU-SA and TPTU-OA.
Since the official implementation of TPTU is not available, we implemented TPTU-SA and TPTU-OA based on the prompts provided in the original paper.
As can be seen in the table, the results clearly demonstrate \OURS's latency and accuracy benefit over both TPTU-SA and TPTU-OA.
Compared with TPTU-SA, \OURS exhibits a significant accuracy improvement due to TPTU’s prevalent issue with repetitive function calls. 
Note that this issue is not fully mitigated even with better prompting (TPTU-SA$^\dagger$), leading to $\sim$15\% of examples failing with repetitive function calls. 
Compared with both TPTU-SA and TPTU-OA, \OURS also benefits from reduced latency through parallel task execution.
Overall, the results are consistent with the main experiments and analysis against other baseline methods (i.e., ReAct and OpenAI's parallel function calling).

\section{User-Supplied Examples for \OURS Configuration}
\label{appendix:prompts}
\OURS provides a simple interface that allows for tailoring the framework to different use cases by providing tool definitions as well as optional in-context examples for the Planner.
Below, we provide the Planner example prompts that are used to set up the framework for the Movie Recommendation and Game of 24 benchmarks with only a few lines of prompts.

\subsection{Movie Recommendation Example Prompts}

\begin{boxA}
\texttt{Question: Find a movie similar to Mission Impossible, The Silence of the Lambs, American Beauty, Star Wars Episode IV - A New Hope}

\texttt{Options:}

\texttt{Austin Powers International Man of Mystery}

\texttt{Alesha Popvich and Tugarin the Dragon}

\texttt{In Cold Blood}

\texttt{Rosetta}
\\
\\
\texttt{1. search("Mission Impossible")}

\texttt{2. search("The Silence of the Lambs")}

\texttt{3. search("American Beauty")}

\texttt{4. search("Star Wars Episode IV - A New Hope")}

\texttt{5. search("Austin Powers International Man of Mystery")}

\texttt{6. search("Alesha Popvich and Tugarin the Dragon")}

\texttt{7. search("In Cold Blood")}

\texttt{8. search("Rosetta")}

\texttt{Thought: I can answer the question now.}

\texttt{9. finish()}

\textbf{\#\#\#}
\end{boxA}

\subsection{Game of 24 Example Prompts}

 \begin{boxA}
\texttt{Question: "1 2 3 4", state\_list: [""]}

\texttt{\$1 = thought\_proposer("1 2 3 4", "")}
\\
\texttt{\$2 = state\_evaluator("1 2 3 4", "\$1")}
\\
\texttt{\$3 = top\_k\_select("1 2 3 4", ["\$1"], ["\$2"])}
\\
\texttt{\$4 = finish()}

\texttt{\#\#\#}

\texttt{Question: "1 2 3 4", state\_list: ["1+2=3(left:3 3 4)","2-1=1(left:1 3 4)","3-1=2(left:2 2 4)","4-1=3(left:2 3 3)","2*1=2(left:2 3 4)"]}
\\
\texttt{\$1 = thought\_proposer("1 2 3 4", "1+2=3(left:3 3 4)")}
\\
\texttt{\$2 = thought\_proposer("1 2 3 4", "2-1=1(left:1 3 4)")}
\\
\texttt{\$3 = thought\_proposer("1 2 3 4", "3-1=2(left:2 2 4)")}
\\
\texttt{\$4 = thought\_proposer("1 2 3 4", "4-1=3(left:2 3 3)")}
\\
\texttt{\$5 = thought\_proposer("1 2 3 4", "2*1=2(left:2 3 4)")}
\\
\texttt{\$6 = state\_evaluator("1 2 3 4", "\$1")}
\\
\texttt{\$7 = state\_evaluator("1 2 3 4", "\$2")}
\\
\texttt{\$8 = state\_evaluator("1 2 3 4", "\$3")}
\\
\texttt{\$9 = state\_evaluator("1 2 3 4", "\$4")}
\\
\texttt{\$10 = state\_evaluator("1 2 3 4", "\$5")}
\\
\texttt{\$11 = top\_k\_select("1 2 3 4", ["\$1", "\$2", "\$3", "\$4", "\$5"], ["\$6", "\$7", "\$8", "\$9", "\$10"])}
\\
\texttt{\$12 = finish()}
\\
\textbf{\#\#\#}

\end{boxA}

\section{Pre-defined \OURS Planner Prompts}
\label{appendix:planner-prompt}

The pre-defined \OURS Planner prompt provides it with specific instructions on how to break down tasks and generate dependency graphs while ensuring that the associated syntax is formatted correctly.
This prompt contains specific rules such as assigning each task to a new line, 
beginning each task with a numerical identifier, and using the \texttt{\$} sign to denote intermediate variables. 

 \begin{boxA}
\texttt{- Each action described above contains input/output types and descriptions.}

\texttt{- You must strictly adhere to the input and output types for each action.}

\texttt{- The action descriptions contain the guidelines. You MUST strictly follow those guidelines when you use the actions.}

\texttt{- Each action in the plan should strictly be one of the above types. Follow the Python conventions for each action.}

\texttt{- Each action MUST have a unique ID, which is strictly increasing.}

\texttt{- Inputs for actions can either be constants or outputs from preceding actions. In the latter case, use the format \$id to denote the ID of the previous action whose output will be the input.}

\texttt{- Ensure the plan maximizes parallelizability.}

\texttt{- Only use the provided action types. If a query cannot be addressed using these, invoke the finish action for the next steps.}

\texttt{- Never explain the plan with comments (e.g. \#).}

\texttt{- Never introduce new actions other than the ones provided.}

\end{boxA}

In addition to user-provided functions, the Planner includes a special, hard-coded \texttt{finish} function.
The Planner uses this function either when the plan is sufficient to address the user query or when it can no longer proceed with planning before executing the current plan, i.e., when it deems replanning necessary.
When the Planner outputs the \texttt{finish} function, its plan generation stops.
Refer to Appendix~\ref{appendix:prompts}  for examples of the Planner's usage of the \texttt{finish} function in planning.
The definition of the \texttt{finish} function is as below and is included as a prompt to the Planner along with the definitions of other user-provided functions.

 \begin{boxA}

\texttt{finish():}

\texttt{  - Collects and combines results from prior actions.}

\texttt{  - A LLM agent is called upon invoking join to either finalize the user query or wait until the plans are executed.}

\texttt{  - join should always be the last action in the plan, and will be called in two scenarios:}

\texttt{    (a) if the answer can be determined by gathering the outputs from tasks to generate the final response.}

\texttt{    (b) if the answer cannot be determined in the planning phase before you execute the plans.}

\end{boxA}

\section{\BENCH Benchmark Generation}
\label{appendix:custom-benchmark}

 Inspired by the IfQA benchmark~\cite{yu2023ifqa}, our custom
benchmark \BENCH contains 113 examples that are designed to use mathematical questions on factual details of different entities to answer questions, thus requiring a mix of search and mathematical operations that are interdependent in various ways. 
For instance, the benchmark includes examples like ``If Texas and Florida were to merge and become one state, as well as California and Michigan, what would be the largest population density among these 2 new states?''  requires four parallel search tasks, followed by math tasks dependent on the search outcomes, that
can be executed in parallel. 

The main objective of the benchmark is to quantify the framework's ability to decompose an input into multiple tasks to derive an answer. 
Therefore, we have meticulously selected 56 distinct entities across various domains whose attributes can be accessible from Wikipedia search. 
By minimizing tool execution (i.e., Wikipedia search) failures, we have aimed our benchmark to effectively assess the frameworks' abilities to decompose questions into multiple tasks, plan them out, and derive final answers based on observations.
Furthermore, to incorporate diverse execution patterns, we crafted various dependency patterns that perform unary and binary math operations after searching for additional information about entities in a given question.
We have also curated different questions that accommodate different numbers of maximally parallelizable tasks, ranging from 2 to 5, and we have included varying numbers of joins between parallel function calls as well to increase problem complexity.
For instance, we have 2 and 3 joins in~\fref{fig:parallel-patterns} (b) and (c), respectively.
The benchmark contains 113 different examples, that were populated by GPT-4 based on the aforementioned criteria and labeled by humans~afterward.

\section{Details of the Game of 24 and the Tree-of-Thoughts Approach}
\label{appendix:tot-details}

The Game of 24 is a mathematical reasoning game that challenges players to manipulate a given set of four numbers, using the basic arithmetic operations of addition, subtraction, multiplication, and division, to arrive at the number 24. The rule of this game is that the given numbers must be used only once. 
For instance, given the numbers 2, 4, 4, and 7, one possible solution is $4 \times (7 - 4) \times 2 = 24$. 
This is a non-trivial reasoning benchmark for LLMs, highlighted by the fact that even advanced models like GPT-4 exhibit only a 4\% success rate, even when using chain-of-thought prompting~\cite{yao2023tree}. 

\begin{figure*}[!t]
\vspace{-2mm}
\centering
\includegraphics[width=0.9\linewidth]{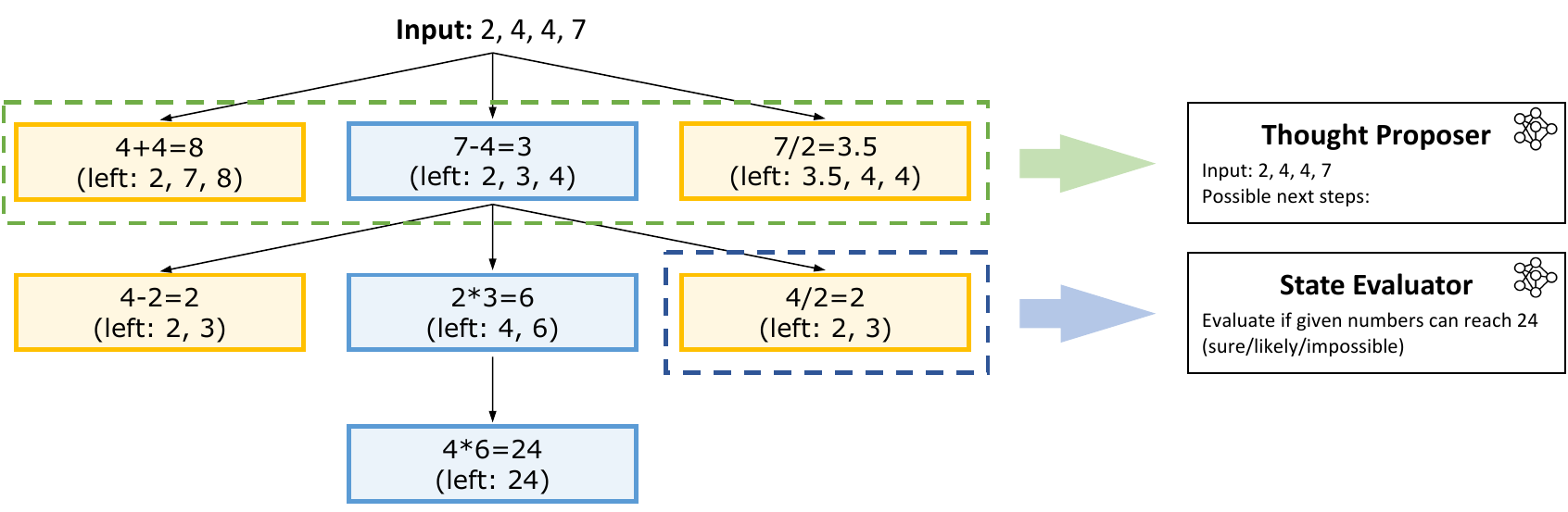}
\caption{Visualization of the Tree of Thoughts (ToT) in the Game of 24. Each node represents a distinct proposal, beginning with the root node and branching out through the application of single operations by the thought proposer. Subsequent states are evaluated by the state evaluator for their potential to reach the target number 24. The ToT retains the top-5 states according to their values.}
  \label{fig:tot_game_24}
\end{figure*}

In ToT, the problem is solved in several steps. At each step, the LLM, referred to as the thought proposer, generates thoughts. Each thought is a partial solution that consists of two numbers and an arithmetic operation between them. 
Then, these
thoughts are fed into the state evaluator which assigns a label for each of them.
These labels are `sure,' `likely,' and `impossible,' which are given to thoughts to denote how
likely they could produce 24 with additional arithmetic operations between the result and the remaining numbers.
Only the thoughts that are likely to produce 24 continue onto the next step.
This process is illustrated in Figure~\ref{fig:tot_game_24}.


\section{Details of WebShop Experiments}
\label{appendix:webshop-details}

\subsection{WebShop Environment}
The WebShop environment simulates an online shopping platform. Tasks are designed for the agent to find the item that best matches the given instruction.
For instance, if the instruction specifies, ``I am looking for a queen-sized bed that is black, and priced lower than 140.00 dollars," 
the agent's task is to pinpoint the bed that precisely fits these criteria: ``queen-sized," ``black," and ``priced under 140.00 dollars."
For each item, there is an associated reward measuring how well this item matches the instruction based on price, item options, and other details contained in the item page. The evaluation metrics are the success rate—the proportion of episodes where the selected product satisfies all requirements—and the average score—the mean reward across episodes. 

\subsection{Baseline Methods}

In addition to ReAct, we use LASER~\cite{ma2023laser} and LATS~\cite{zhou2023language}  as baseline methods to compare against \OURS.
LASER~\cite{ma2023laser} solves tasks through a state-exploration approach. In the context of WebShop, the possible environment pages are encoded as different states (e.g., search page, item page, and item detail subpage). Actions are used to transition between these states, such as executing a search query, selecting an item, checking the item detail, navigating the next search page and so on. The Webshop exploration is therefore reduced to a search problem on the given state-space graph.

Using a variant of Monte Carlo Tree Search, LATS~\cite{zhou2023language} plans its actions by constructing a decision tree, evaluating potential moves based on their likelihood of success, and selecting actions through a balance of exploration and exploitation. The agent then adapts its strategy based on feedback from the environment, learning from both successes and failures to refine its decision-making process. This iterative approach allows LATS to navigate complex online shopping tasks, albeit much more slowly due to its exhaustive tree search.


\end{document}